\documentclass[onecolumn,journal]{IEEEtran}
\usepackage[T1]{fontenc}
\usepackage{amsmath, amsfonts, amsthm, amssymb}
\usepackage{algorithmic}
\usepackage{algorithm}
\usepackage{array}
\usepackage[caption=false,font=normalsize,labelfont=sf,textfont=sf]{subfig}
\usepackage{textcomp}
\usepackage{stfloats}
\usepackage{url}
\usepackage{verbatim}
\usepackage{graphicx}
\usepackage{cite}
\usepackage{bm} 
\usepackage{arydshln} 
\usepackage{xcolor}
\usepackage[colorlinks,linkcolor=red,anchorcolor=blue,citecolor=brown]{hyperref}
\usepackage{multirow}

\newtheorem{rem}{Remark}
\newtheorem{ass}{Assumption}

\begin{document}

\title{Observability Analysis and Composite Disturbance Filtering for a Bar Tethered to Dual UAVs Subject to Multi-source Disturbances}

\author{Lidan Xu, Dadong Fan, Junhong Wang, Wenshuo Li, Hao Lu*, and Jianzhong Qiao*
\thanks{This paper was produced by the IEEE Publication Technology Group. They are in Piscataway, NJ.}
\thanks{Manuscript received ; revised .}
\thanks{Lidan Xu is with the School of Cyber Science and Technology, Beihang University, Beijing, 100191, China.}
\thanks{Dadong Fan, Junhong Wang, and Jianzhong Qiao are with the School of Automation Science and Electrical Engineering, Beihang University, Beijing, 100191, China.}
\thanks{Hao Lu is with Beijing Engineering Research Center
 of Industrial Spectrum Imaging, School of Automation and Electrical Engineering, University of Science and Technology Beijing, Beijing 100083, China.}
\thanks{Wenshuo Li is with Hangzhou Innovation Institute, Beihang University, Zhejiang, 310052, China.}
}

\maketitle

\begin{abstract}
    Cooperative suspended aerial transportation is  highly susceptible to multi-source disturbances such as aerodynamic effects and thrust uncertainties.
    To achieve precise load manipulation,
    existing methods often rely on extra sensors to measure cable directions or the payload's pose,
    which increases the system cost and complexity. 
    A fundamental question remains: is the payload's pose observable under multi-source disturbances using only the drones' odometry information?
    To answer this question, this work focuses on the two-drone-bar system and proves that the whole system is observable when only two or fewer types of lumped disturbances exist by using the observability rank criterion.
    To the best of our knowledge, we are the first to present such a conclusion
    and this result paves the way for more cost-effective and robust systems by minimizing their sensor suites.
    Next, to validate this analysis,
    we consider the situation where the disturbances are only exerted on the drones,
    and develop a composite disturbance filtering scheme.
    A disturbance observer-based  error-state extended Kalman filter is designed for both state and disturbance estimation,
    which renders improved estimation performance for the whole system evolving on the manifold $(\mathbb R^3)^2 \times (\mathsf{TS}^2)^3$.
    Our simulation and experimental tests have validated that
    it is possible to fully estimate the state and disturbance of the system with only odometry information of the drones.    
\end{abstract}

\section{Introduction}
\subsection{Motivation}
 Unmanned Aerial Vehicles (UAVs) have received much attention in the last decade due to their great potential for complex tasks, ranging from search and rescue to load manipulation and transportation.
 Compared with ground transportation,
 aerial transportation can overcome terrain obstacles and traffic congestion,
 which provides a faster and more energy-efficient solution. 
 This research adopts the cable connection structure since it is lightweight and preserves the agility of the UAV \cite{zeng2020differential}.
 
 Compared with the single-drone transportation,
 multi-drone transportation can provide higher load capacity, lower cost, extra payload maneuverability \cite{sreenath2013dynamics}, and more robustness to vehicle failure \cite{liang2021fault}.
 The state-of-the-art methods for control and state estimation of the aerial cooperative transportation systems can be classified into two groups: UAV-based design and payload-based design.
 In the UAV-based design,
 the effect of the payload on the drones is treated as external disturbance that has either a known value or known bounds \cite{geng2020cooperative}.
 Instead of simply closing the loop on UAVs, payload-based design requires real-time pose feedback of the payload, where efforts are mainly focused on payload dynamics, and UAVs act as actuators driving the payload to the desired states \cite{lee2017geometric}.
 
 This research focuses on the payload-based design due to its tremendous potential in the complex task of load pose manipulation.
 However, disturbances exerted on the UAVs (e.g., aerodynamic effects and thrust uncertainty) and the payload (e.g., wind disturbances, mass uncertainty) hinder the precise modeling and refined analysis of the whole transportation system. 
 Furthermore, 
 its practical implementation is severely hampered by a critical dependency: the need for accurate, real-time payload pose information. Existing solutions largely rely on installing extra sensors on the payload itself, which introduces significant limitations. This fundamental limitation motivates our research into a novel estimation paradigm that eliminates the need for payload-mounted sensors, relying solely on mature and readily available UAV odometry data.

 \subsection{Related Works}
 For the single-drone-payload case, 
 a single load cell is attached to the slung load to measure the cable tension,
 which is combined with the onboard inertial measurement unit (IMU) to estimate the swing angle \cite{lee2017autonomous}.
 In \cite{klausen2017nonlinear},
 the displacement angles of the slung load are measured by two magnetic encoders and used through a delayed feedback approach to reduce the swing motion of the load.
 For the multi-drone-payload case, the most common method to measure the payload pose is through the motion capture system, which can capture the load pose with a sampling rate of up to 200 Hz, but it is limited in the indoor environment.
 For the outdoor scenario,
 a GPS+compass group is installed on the payload to provide real-time state information, which is combined with the desired trajectory of the payload to optimize the cable attachment geometry online \cite{geng2020cooperative}.
 Instead of using an external sensing system, a cooperative estimation scheme is proposed in \cite{li2021cooperative} to infer the payload’s full 6-DoF states by using the downward camera and IMU onboard.
 In this approach, the attachment points on the payload are captured by the cameras, 
 and the payload pose estimation is subsequently solved by sharing the drones’ local position estimates and their relative positions with respect to the payload.
 This approach is verified experimentally in an indoor setting, but it is limited to situations where load swing is small so that the load attachment points are always within the camera view.
 In a payload-swing scenario, an IMU is placed on the cargo to feedback the spatial swing angle of the cargo in real time \cite{payloadIMU}.
 Based on the swing angle feedback, a nonlinear adaptive controller is proposed to enhance cargo swing damping while considering the influence of unknown air resistance on quadrotors and cargo during transportation.
 Although the above sensor suites provide a promising solution for payload sensing,
 the installation of extra sensors increases the structural weight and complexity of the overall system,
 and is highly dependent on customized designs.
 
 In contrast to the payload sensing technology, multi-UAV localization technology is more mature.
 Onboard IMUs are popularly combined with other sensors such as GPS \cite{GPSRTK}, cameras \cite{d2slam}, and light detection and ranging sensors (LiDARs) \cite{swarmlio2} to provide accurate and robust pose estimation for drone swarms in the outdoor environment.
 Using only odometry information of the drones, pose estimation of the payload is being investigated.
 For the single-drone-payload case,
 a state observer is established in \cite{wang2022geometric} to estimate the linearized swing angle of the cable for precise trajectory tracking of the drone.
 A filtering technique based on a set of recursive equations is proposed in \cite{de2019swing} to estimate the swing angle only using the data available from the onboard IMU.
 The authors further develop an improved method in \cite{de2023improved} to estimate the swing angle based on the extended Kalman filter (EKF) structure, which considers the additional aerodynamic disturbance force and allows for improved estimation accuracy in both stationary and maneuvering flight. 
 In addition, a neural network estimator is proposed in \cite{mellet2023neural} to perform real-time payload position estimation, where the neural network is trained with domain randomization and shows accurate zero-shot estimation even with excitations never seen by the system before.
 For the multi-drone-payload case,
 relevant literature is quite rare.
 In \cite{xie2020}, a lightweight estimator based on an EKF is designed for a simplified system composed by two aerial robots and a tethered point mass,
 where the acceleration of the leader robot is neglected,
 and the follower stabilizes the system with respect to the leader using only feedback from its IMU.
 Furthermore, a numerical observability analysis is conducted in \cite{xie2020},
 but it does not consider any uncertainties in the system. 
 A less conservative solution for the payload pose estimation is proposed in \cite{xu2024oscillation} for a bar-shaped payload suspended by two drones,
 where a refined cooperative disturbance estimation strategy is developed based on the linearized model to capture the pose dynamics of the payload and the model uncertainties on the drones, and the estimates are then utilized for payload stabilization.
 Although the above methods provide a filter solution for pose estimation of the suspended payload,
 they do not consider the full observability of the payload under multi-source disturbances,
 which may degrade the state estimation accuracy and ultimately worsen the control performance. 
 In summary, the literature reveals a clear gap: a general and robust solution for estimating the pose of a payload transported by multiple UAVs under multi-source disturbances, without relying on payload-mounted sensors, is still lacking. 
 To the best of our knowledge, the fundamental observability of the system under such conditions remains an open question. Without a clear understanding of observability, the performance and convergence guarantee of any proposed filters are questionable. This gap severely limits the practical application for multi-UAV transportation in complex environments.

 In terms of filtering methods, Kalman filter and its variants have been widely used in modern control systems.
 However, Kalman filters typically treat the state space as a Euclidean space $\mathbb R^n$ while many practical systems (e.g., robotic systems) usually have their states evolving on manifolds (e.g., rotation group $\mathsf{SO}(3)$).
 To circumvent the constraints imposed by the manifold, an
 elegant and effective way is to perform Kalman filtering steps
 (i.e., predict and update) on the error-state, i.e., the ESEKF, which has been widely used in attitude estimation \cite{markley2003attitude}, online extrinsic calibration \cite{mirzaei2008kalman}, GPS/IMU navigation \cite{sola2017quaternion}, visual inertial navigation \cite{li2013high} and LiDAR inertial navigation \cite{xu2021fast}.
 The basic idea of ESEKF is to repeatedly parameterize the state trajectory on the manifold by an error state
 trajectory in the vector space, perform a normal EKF on the error state trajectory to
 update the error state, and add it back
 to the original state on manifolds to obtain the posterior estimate.
 Since this error is small, this allows us to work with local parameterizations of the manifold around each point instead of a singular global minimal parametrization (e.g. Euler angles for $\mathsf{SO}(3)$, polar and azimuthal angles for $\mathsf{S}^2$).
 Despite of these advantages,
 when the system is subject to various disturbances from multiple sources,
 accurate state estimates can hardly be derived by utilizing the existing ESEKF methods.
 The key to precise state estimation is the refined modelling and separation of these multi-source disturbances \cite{guo2012initial,guo2013anti}.
 Following this idea, the composite disturbance filtering (CDF) scheme has been put forward to address the state estimation problem for systems with multi-source heterogeneous disturbances \cite{guo2014anti}.
 Specifically, the CDF is a refined filtering methodology, which aims at simultaneous disturbance rejection, attenuation, and absorption via a DO-based composite hierarchical anti-disturbance estimation architecture \cite{guo2023composite}.
 This inspires us to combine the ESEKF with the disturbance observer (DO) for improved estimation of the aerial co-transportation system within the CDF framework .

 \subsection{Contributions}
  In this paper, we focus on the two-drone-bar case and analyze the observability of the payload pose under multiple disturbances in depth. 
  Then, a DO-based ESEKF method is developed for both state and disturbance estimation of the whole system.
  The key contributions of our work are as follows:
 \begin{itemize}
  \item With only odometry information of the drones, observability for the payload pose under multi-source disturbances is analyzed in depth for the first time by using the observability rank criterion (ORC),
  which reveals that the payload pose is observable under mild conditions if only two or fewer types of lumped disturbances are exerted on the overall system,
  thereby providing a theoretical foundation for filter design in such complex scenarios.
  \item Building upon the observability analysis, we propose a DO-based ESEKF for both state and disturbance estimation of the whole transportation system.
  This on-manifold nonlinear filter for estimation of the payload pose is proposed for the first time,
  overcoming the limitations of linearized models (e.g., \cite{xu2024oscillation}) and enabling accurate estimation far from the equilibrium.
  \item  Extensive numerical simulations are conducted to validate the effectiveness of our proposed approach and better estimation performance is demonstrated compared with the linearized case in \cite{xu2024oscillation}.
  Furthermore, experimental results validate the practicality of our proposed filter in engineering practice,
  paving the way for its application in robust payload-based control without extra sensors.
\end{itemize}
 
\section{Problem Formulation}
 In this section, the kinematic and dynamic model describing the multibody system are derived.
 Then, the multi-source disturbances exerted on the system are analyzed and summarized into the lumped form.
 Finally, the system observations are presented.

 \subsection{Nominal Model}
\begin{table}[!ht]\footnotesize
	\renewcommand\arraystretch{1.5}
	\caption{Nomenclature} 
	\centering
	\begin{tabular}{lp{6cm}} 
		\hline
		 \textbf{Symbols} & \textbf{Physical interpretations} \\ 
         \hline
        $\bm d_{\bm p_0} \in \mathbb R^3$ & translational disturbance on the payload\\
		$\bm d_{\bm p_i} \in \mathbb R^3$ & translational disturbance on the $i$th drone\\
        $\bm d_{\bm q_0} \in \mathbb R^3$ & rotational disturbance on the payload\\
		 $\mathcal F^{\mathcal I}, \mathcal F^{\mathcal B_i}$ & inertial frame, $i$th drone frame\\
		$g \in \mathbb{R}^{+}$ & gravity constant\\
            $J_0 \in \mathbb R^+$ & rotational inertia of the bar-shaped payload\\
		$l_i \in \mathbb{R}^{+}$ &  length of the $i$th  cable\\
		$ m_0, m_i \in \mathbb{R}^{+} $ & mass of the payload, $i$th drone\\ 
	 $\bm{p}_0, \bm p_i \in \mathbb{R}^3$ & CoM position of the payload, $i$th drone\\
		$\bm{v}_0, \bm v_i \in \mathbb{R}^3$ & CoM velocity of the payload, $i$th drone\\
        $\dot{\bm{v}}_0, \dot{\bm v}_i \in \mathbb{R}^3$ & CoM acceleration of the payload, $i$th drone\\
        $\bm q_0 \in \mathsf S^2$ & direction of the bar axis\\
		 $\bm q_i \in \mathsf S^2$ & direction of the $i$th cable\\
		$\bm u_i \in \mathbb R^3$ & desired thrust for the $i$th drone\\
		$\bm \mu_i \in \mathbb{R}^3$ & tension of the cable on the $i$th drone\\
        $\bm \omega_0 \in \mathsf T_{\bm q_0} \mathsf S^2$ &  angular velocity of the payload\\
		$\bm \omega_i \in \mathsf T_{\bm q_i} \mathsf S^2$ &  angular velocity of the $i$th cable\\
		\hline
	\end{tabular} 
	\label{symbols}
\end{table}
\begin{figure}
    \centering
    \includegraphics[width=3.5in]{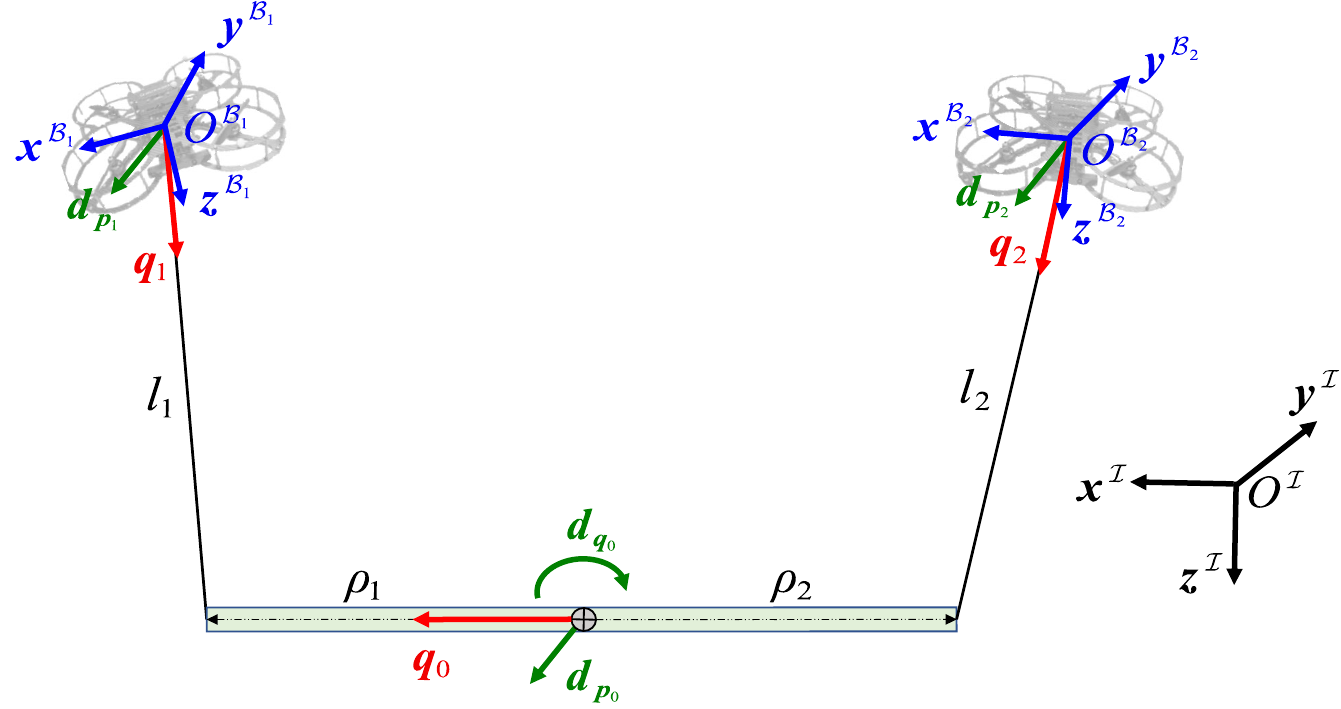}
    \caption{Schematic of the two-drone-bar system}
    \label{shiyitu}
\end{figure}
Consider two aerial vehicles carrying a bar-shaped payload through massless cables of fixed lengths as shown in Fig. \ref{shiyitu}.
The cables are assumed taut and attached to the center of mass (CoM) of the drones,
 and therefore have no effects on the rotation of the drones.
 The unit vectors are defined as $\bm e_1 = [1,0,0]^\top$, $\bm e_2 = [0,1,0]^\top$, and $\bm e_3 = [0,0,1]^\top$.
 We choose the north-east-down (NED) frame as the inertial frame $\mathcal F^{\mathcal I} = \{O^{\mathcal I}, \bm x^{\mathcal I}, \bm y^{\mathcal I}, \bm z^{\mathcal I}\}$, with $O^{\mathcal I}$ fixed at a point on the ground.
 The body-attached frame for the $i$th drone is denoted by $\mathcal F^{\mathcal B_i} = \{O^{\mathcal B_i}, \bm x^{\mathcal B_i}, \bm y^{\mathcal B_i}, \bm z^{\mathcal B_i}\}$ as shown in Fig. \ref{shiyitu}.  
Throughout this paper, the variables related to the payload are denoted by the subscript 0, and the variables for the $i$th drone are denoted by the subscript $i \in \{1,2,3\}$.
Some frequently used symbols are provided in Table \ref{symbols}.
 Since the bar-shaped payload is treated as a one-dimension rigid body, which has no rotation freedom around the bar axis \cite{pereira2020pose},
 the rotational inertia of the payload is isotropic and therefore defined as the scalar $J_0 \in \mathbb R^+$ as shown in Table \ref{symbols}.

The configuration space of the complete multibody system is given by 
\begin{equation*}
	\mathsf Q = \underbrace{\mathbb R^3 \times \mathsf S^2}_{\text{Bar-shaped load pose}} \times \underbrace{(\mathsf{S}^2)^2}_{\text{Cable directions}} \times  \underbrace{(\mathsf{SO}(3))^2}_{\text{Drone attitudes}},
\end{equation*}
where $\mathsf{SO}(3)$ denotes the special orthogonal group of $3 \times 3$ rotation matrices,
 and $\mathsf{S}^2$ denotes the two-sphere group of $3 \times 1$ unit vectors.
 
  Since the attitude dynamics of the drones is assumed to be sufficiently fast and accurate to track the desired attitude command,
  the desired thrust $\bm u_{i} \in \mathbb R^3 \ (i=1,2)$ of the $i$th drone is directly treated as the control input for the multibody system and the attitude loops of the drones are omitted for simplicity,
  which is common practice in aerial transportation community \cite{klausen2018cooperative,lee2017geometric,erskine2019wrench}.
 Then the state for describing the system configuration and its velocity is defined by $(\bm p_0, \bm v_0, \bm q_0, \bm \omega_0, \bm q_1, \bm \omega_1, \bm q_2, \bm \omega_2)$, which evolves on the Cartesian product manifold $(\mathbb R^3)^2 \times (\mathsf{TS}^2)^3$.
 The system kinematics is defined as 
 \begin{equation}\label{kinematics}
    \begin{aligned}
        \dot{\bm p}_0 &= \bm v_0 \\
        \dot{\bm q}_0 &= \bm \omega_0^\times \bm q_0\\
        \dot{\bm q}_1 &= \bm \omega_1^\times \bm q_1\\
        \dot{\bm q}_2 &= \bm \omega_2^\times \bm q_2
    \end{aligned}
\end{equation}
where the notation $(\cdot)^\times$ is the skew-symmetric cross product matrix of the vector $(\cdot) \in \mathbb R^3$.
 The dynamic model of the multibody system is derived from Lagrange D’Alembert variational principle as follows (see \cite{goodman2022geometric} for details) 
 \begin{equation}\label{nominaldynamics}
	\begin{aligned}
		&m_T \dot{\bm{v}}_0 + \sum_{i=1}^{2} m_i [(-1)^i \rho_i (\bm q_0^{\times} \dot{\bm{\omega}}_0 + \|\bm{\omega}_0\|^2  \bm q_0) + l_i (\bm{q}_i^{\times} \dot{\bm{\omega}}_i  + \|\bm{\omega}_i\|^2  \bm q_i)]
        = m_T g \bm{e}_3 + \sum_{i=1}^{2} \bm u_i,\\
		&\bar{J}_0 \dot{\bm{\omega}}_0 + \sum_{i=1}^{2} (-1)^{i+1} m_i \rho_i \bm q_0^\times (\dot{\bm{v}}_0 + l_i \bm{q}_i^{\times} \dot{\bm{\omega}}_i + l_i \| \bm{\omega}_i \|^2 \bm{q}_i) = \sum_{i=1}^{2} (-1)^{i+1} \rho_i \bm q_0^\times (\bm u_i + m_i g \bm{e}_3),\\
		&m_i l_i \dot{\bm{\omega}}_i - m_i \bm{q}_i^{\times} [\dot{\bm{v}}_0 + (-1)^{i} \rho_i \bm q_0^{\times} \dot{\bm{\omega}}_0 - \rho_i \|\bm \omega_0\|^2 \bm{q}_0] = - \bm{q}_i^{\times} (\bm u_i + m_i g \bm{e}_3), \quad i = 1, 2,
	\end{aligned}
\end{equation}
where $m_T = \sum\limits_{i=0}^2{m_i}$ and $\bar J_0 = J_0 +\sum\limits_{i=1}^2{m_i \rho_i^2}$ are used for substitutions.

\subsection{Disturbance Analysis}
According to the sources, the multi-source disturbances on the overall system can be classified into two groups: disturbances on the drones and disturbances on the payload.

The disturbances on the drones mainly include the thrust uncertainty and aerodynamic effects between the drones, which affect the overall system dynamics through the drones' accelerations.
Therefore, we denote the disturbances on the drones as $\bm d_{\bm p_i} \ (i=1,2)$,
which are exerted on the translational channel of the $i$th vehicle.
The input matrices of the disturbances $\bm d_{\bm p_i} \ (i=1,2)$ are the same as those of the control inputs $\bm u_i \ (i=1,2)$, respectively.

The disturbances on the bar-shaped payload include the mass uncertainty, rotational inertia uncertainty, wind disturbance, etc,
which have direct effects on the payload pose dynamics.
Therefore, we denote the disturbances on the payload as $\bm d_{\bm p_0}$ and $\bm d_{\bm q_0}$,
which are treated as perturbations on the translational acceleration $\dot{\bm v}_0$ and angular acceleration $\dot{\bm \omega}_0$ of the payload, respectively.
Since the bar-shaped payload does not have rotation freedom around the bar axis,
the disturbance $\bm d_{\bm q_0}$ is always perpendicular to the direction of the bar axis $\bm q_0$.

To summarize, there are four types of lumped disturbances exerted on the overall system: $\bm d_{\bm p_0}$, $\bm d_{\bm q_0}$, $\bm d_{\bm p_1}$, and $\bm d_{\bm p_2}$, which enter the system through different channels.
The dynamic model subject to these multi-source disturbances is given as
 \begin{subequations}\label{fulldynamics}
	\begin{align}
		&m_T \dot{\bm{v}}_0 + \sum_{i=1}^{2} m_i [(-1)^i \rho_i (\bm q_0^{\times} \dot{\bm{\omega}}_0 + \|\bm{\omega}_0\|^2  \bm q_0)+ l_i \bm{q}_i^{\times} \dot{\bm{\omega}}_i + l_i \|\bm{\omega}_i\|^2  \bm q_i]
        = m_T g \bm{e}_3 + \sum_{i=1}^{2} (\bm u_i + \bm d_{\bm p_i}) + \bm d_{\bm p_0},\label{translation}\\
		&\bar{J}_0 \dot{\bm{\omega}}_0 + \sum_{i=1}^{2} (-1)^{i+1} m_i \rho_i \bm q_0^\times (\dot{\bm{v}}_0 + l_i \bm{q}_i^{\times} \dot{\bm{\omega}}_i + l_i \| \bm{\omega}_i \|^2 \bm{q}_i) = \sum_{i=1}^{2} (-1)^{i+1} \rho_i \bm q_0^\times (\bm u_i + \bm d_{\bm p_i} + m_i g \bm{e}_3) + \bm q_0^\times \bm d_{\bm q_0},\label{rotation}\\
		&m_i l_i \dot{\bm{\omega}}_i - m_i \bm{q}_i^{\times} [\dot{\bm{v}}_0 + (-1)^{i} \rho_i \bm q_0^{\times} \dot{\bm{\omega}}_0 - \rho_i \|\bm \omega_0\|^2 \bm{q}_0] = - \bm{q}_i^{\times} (\bm u_i + \bm d_{\bm p_i} + m_i g \bm{e}_3), \quad i = 1, 2,\label{cable_direction}
	\end{align}
\end{subequations}
\subsection{Problem Statement}
For the system \eqref{fulldynamics}, there are three types of observations, the positions of the drones $\bm p_i \ (i=1,2)$, the manifold constraints $(\bm q_i, \bm \omega_i) \in \mathsf{TS}^2$, and the 
orthogonal constraint $\bm d_{\bm q_0} \perp \bm q_0$
\begin{equation}\label{measurements}
    \begin{aligned}
        \bm h_1 &=
        \begin{bmatrix}
            \bm p_1\\
            \bm p_2
        \end{bmatrix}
        =
        \begin{bmatrix}
            \bm p_0 + \rho_1 \bm q_0 - l_1 \bm q_1\\
            \bm p_0 - \rho_2 \bm q_0 - l_2 \bm q_2
        \end{bmatrix}\\
        \bm h_2 &=
        \begin{bmatrix}
            1\\
            0
        \end{bmatrix}
        \ =
        \begin{bmatrix}
            \bm q_i^\top \bm q_i\\
            \bm q_i^\top \bm \omega_i
        \end{bmatrix}, \quad i=0,1,2\\
        \bm h_3 &=
        \ 0
        \quad =
        \bm q_0^\top \bm d_{\bm q_0}
    \end{aligned}
\end{equation}
Though the velocities and the accelerations of the drones may also be measured by the onboard sensors,
these measurements are omitted since they have been included in the high-order time derivatives of the output $\bm h_1$.
With the measurements in \eqref{measurements}, it is uncertain whether the state of the model in \eqref{fulldynamics} is observable subject to multi-source disturbances $(\bm d_{\bm p_0}, \bm d_{\bm q_0}, \bm d_{\bm p_1}, \bm d_{\bm p_2})$,
which will be answered in the subsequent section.

\section{Observability Analysis}\label{observability}
 In this section, the original model is analyzed and transformed into a reduced form for observability test.
 By making assumptions on the multi-source disturbances and system dynamics, we prove that the payload pose is observable under mild conditions if only two or fewer types of lumped disturbances exist.
 
\subsection{Model Fundamentals}
Directly computing the Lie derivatives of the outputs in \eqref{measurements} symbolically to verify the system observability is quite complicated, since the system dynamics \eqref{fulldynamics} involves the $12 \times 12$ matrix inverse. Therefore, we attempt to transform the dynamical system \eqref{fulldynamics} into a reduced form.

The equation \eqref{translation} corresponds to the translational dynamics of the payload
\begin{equation}\label{payloadtrans}
    \begin{aligned}
        m_0 \dot{\bm v}_0 &= -\bm \mu_1 - \bm \mu_2 +\bm{d}_{\bm{p}_0}+m_0 g\bm{e}_3\\
    \end{aligned}
\end{equation}
The equation \eqref{rotation} corresponds to the rotational dynamics of the payload
\begin{equation}\label{payloadrot}
    \begin{aligned}
        J_0 \dot{\bm \omega}_0 &= - \bm q_0^\times (\rho_1 \bm{\mu}_1 - \rho_2 \bm{\mu}_2) + \bm q_0^\times \bm d_{\bm q_0}
    \end{aligned}
\end{equation}
The equation \eqref{cable_direction} corresponds to the direction constraints of the cable forces
\begin{equation}\label{cableforceconstraints}
    \begin{aligned}
        \bm q_i^\times \bm \mu_i = \bm 0_{3\times 1}, \quad i=1,2
    \end{aligned}
\end{equation}
By combining the direction constraints \eqref{cableforceconstraints} and the magnitude constraints of the cable directions $\|\bm q_i\| = 1 \ (i=1,2)$, the cable direction can be replaced by
\begin{equation}\label{cabledirect}
    \begin{aligned}
        \bm q_i = \frac{\bm \mu_i}{\|\bm \mu_i \|}, \quad i=1,2
    \end{aligned}
\end{equation}

Based on the system kinematics \eqref{kinematics} and the equations \eqref{payloadtrans} \eqref{payloadrot} \eqref{cabledirect}, the reduced system model is given as
\begin{equation}
\begin{aligned}
        \dot{\bm p}_0 &= \bm v_0\\
        \dot{\bm v}_0 &= -\frac{1}{m_0} \bm \mu_1 - \frac{1}{m_0} \bm \mu_2 + \frac{1}{m_0} \bm d_{\bm p_0} + g \bm e_3\\
        \dot{\bm q}_0 &= \bm \omega_0^\times \bm q_0\\
        \dot{\bm \omega}_0 &= - \frac{1}{J_0}\bm q_0^\times (\rho_1 \bm{\mu}_1 - \rho_2 \bm{\mu}_2) + \frac{1}{J_0} \bm q_0^\times \bm d_{\bm q_0}
\end{aligned}
\end{equation}
According to the force analysis of the drones, the cable force on the $i$th drone satisfies
\begin{equation}\label{cableforceexpression}
    \begin{aligned}
        \bm \mu_i = m_i \dot{\bm v}_i - \bm u_i - \bm d_{\bm p_i} - m_i g \bm e_3, \quad i=1,2
    \end{aligned}
\end{equation}
where $\dot{\bm v}_i \ (i=1,2)$ is expressed in $\mathcal F^{\mathcal I}$ and can be estimated by the onboard data fusion algorithms.

Since the two-drone-bar system operates under steady conditions most of the time,
the following disturbance assumption is made
\begin{ass}\label{disturbanceAss}
The multi-source disturbances on the two-drone-bar system $(\bm d_{\bm p_0}, \bm d_{\bm q_0}, \bm d_{\bm p_1}, \bm d_{\bm p_2})$ change slowly, i.e.,
\begin{equation}
    \begin{aligned}
        \dot{\bm d}_{\bm p_i} &= \bm 0_{3\times 1}, \quad i=0,1,2\\
        \dot{\bm d}_{\bm q_0} &= \bm 0_{3\times 1}
    \end{aligned}
\end{equation}
\end{ass}
Combining Assumption \ref{disturbanceAss} and the expressions of cable forces \eqref{cableforceexpression},
the reduced model is extended as:
\begin{equation}\label{finalmodel}
\begin{aligned}
    \dot{\bm p}_0 &= \bm v_0\\
    \dot{\bm v}_0 &= -\frac{1}{m_0} \sum_{i=1}^2{\bm u_i^{tot}} + \frac{1}{m_0} \sum_{i=0}^2{\bm{d}_{\bm{p}_i}} + g \bm e_3\\
    \dot{\bm q}_0 &= \bm \omega_0^\times \bm q_0\\
    \dot{\bm \omega}_0 &= \frac{1}{J_0} \sum_{i=1}^2{(-1)^{i} \rho_i \bm q_0^\times (\bm u_i^{tot}-\bm d_{\bm p_i})} + \frac{1}{J_0} \bm q_0^\times \bm d_{\bm q_0} \\
    \dot{\bm d}_{\bm p_0} &= \bm 0_{3 \times 1}\\
    \dot{\bm d}_{\bm q_0} &= \bm 0_{3 \times 1}\\
    \dot{\bm d}_{\bm p_1} &= \bm 0_{3 \times 1}\\
    \dot{\bm d}_{\bm p_2} &= \bm 0_{3 \times 1}\\
\end{aligned}
\end{equation}
where $\bm u_i^{tot} = m_i \dot{\bm v}_i - \bm u_i - m_i g \bm e_3 \ (i=1,2)$ can be treated as the known inputs for the extended system \eqref{finalmodel}.
Therefore, the extended system states are defined as $\bm x = [\bm p_0^\top, \bm v_0^\top, \bm q_0^\top, \bm \omega_0^\top, \bm d_{\bm p_0}^\top, \bm d_{\bm q_0}^\top, \bm d_{\bm p_1}^\top, \bm d_{\bm p_2}^\top]^\top$,
and the inputs defined as
$\bm u = [(\bm u_1^{tot})\top, (\bm u_2^{tot})^\top]^\top$.
By combining the original outputs \eqref{measurements} and the expressions of cable forces \eqref{cableforceexpression},
the reduced observations for the extended system \eqref{finalmodel} are presented as
\begin{equation}\label{finaloutput}
    \begin{aligned}
    \bm h_1 &=
    \begin{bmatrix}
        \bm p_0 + \rho_1 \bm q_0 - l_1 \frac{\bm u_1^{tot}-\bm d_{\bm p_1}}{\|\bm u_1^{tot}-\bm d_{\bm p_1}\|}\\
        \bm p_0 - \rho_2 \bm q_0 - l_2 \frac{\bm u_2^{tot}-\bm d_{\bm p_2}}{\|\bm u_2^{tot}-\bm d_{\bm p_2}\|}\\
    \end{bmatrix}\\
    \bm h_2^{r} &=
        \begin{bmatrix}
            \bm q_0^\top \bm q_0\\
            \bm q_0^\top \bm \omega_0\\
        \end{bmatrix}\\
    \bm h_3 &=
        \begin{bmatrix}
            \bm q_0^\top \bm d_{\bm q_0}
        \end{bmatrix}
    \end{aligned}
\end{equation}
To this point, the reduced nonlinear model \eqref{finalmodel} \eqref{finaloutput} for the subsequent observability analysis is derived. 
\subsection{Linearized Observability Analysis}
We attempt to linearize the reduced system \eqref{finalmodel} \eqref{finaloutput} around its equilibrium and directly analyze the observability of the linearized system.
\begin{rem}
There are two primary reasons for adopting the linearized observability approach.
On one hand, payload transportation typically does not demand high system maneuverability, and therefore the two-drone-bar system can operate near the equilibrium most of the time, for which linearized observability is sufficient.
On the other hand, though the agile motion of the drones can inject additional excitation signals into the system, which might alter the rank of the nonlinear observability matrix \cite{arthur}, 
such excitation is inherently transient and can not be sustained throughout the entire transportation mission.
In contrast, the linearized observability results offer broader applicability for some degenerate motion, such as hover, uniform rectilinear motion, etc. 
\end{rem}

Firstly, the equilibrium for the system \eqref{finalmodel} is chosen as $\bm x_e = [\bm p_{0e}^\top, \bm v_{0e}^\top, \bm q_{0e}^\top, \bm \omega_{0e}^\top, \bm d_{\bm p_0, e}^\top, \bm d_{\bm q_0, e}^\top, \bm d_{\bm p_1, e}^\top, \bm d_{\bm p_2, e}^\top]^\top$\footnote{States, disturbances, and inputs of the chosen equilibrium are represented with a subscript $e$.}.
At this equilibrium, the entire system is assumed to move at a constant speed $\bm v_{0e}$, 
and the attitude of the payload and the cable forces are assumed to be time invariant, i.e., 
\begin{equation}\label{forcebalance}
    \begin{aligned}
        \bm 0_{3\times 1} &= -\frac{1}{m_0} \sum_{i=1}^2{\bm \mu_{ie}} + \frac{1}{m_0} \bm{d}_{\bm{p}_0, e} + g \bm e_3\\
        \bm 0_{3\times 1} &= \bm \omega_{0e}^\times \bm q_{0e}\\
        \bm 0_{3\times 1} &= \frac{1}{J_0} \sum_{i=1}^2{(-1)^{i} \rho_i \bm q_{0e}^\times \bm \mu_{ie}} + \frac{1}{J_0} \bm q_0^\times \bm d_{\bm q_0, e} \\
    \end{aligned}
\end{equation}
where $\bm \mu_{ie} = \bm u_{ie}^{tot} - \bm d_{\bm p_i, e} \ (i=1,2)$ denotes the cable force at the equilibrium.
It can be computed from equations \eqref{forcebalance} that
\begin{equation}\label{cableforceeq}
    \begin{aligned}
    \bm \omega_{0e} &= \bm 0_{3\times 1}\\
    \bm \mu_{1e} &= \frac{\rho_2(m_0 g \bm e_3 + \bm d_{\bm p_0,e})+\bm d_{\bm q_0, e} - n_0 \bm q_{0e}}{\rho_1+\rho_2}\\
    \bm \mu_{2e} &= \frac{\rho_1(m_0 g \bm e_3 + \bm d_{\bm p_0,e}) - \bm d_{\bm q_0, e} + n_0 \bm q_{0e}}{\rho_1+\rho_2}\\
    \end{aligned}
\end{equation}
where $n_0 \in \mathbb R$ characterizes the magnitude of internal force in the bar \cite{xu2024force}.

Since the manifold constraint of the state $(\bm q_0, \bm \omega_0) \in \mathsf{TS}^2$ has been considered in the output \eqref{finaloutput},
jacobian linearization technology is utilized to decouple the system model.
The state offset around the equilibrium is expressed as
\begin{equation}
    \begin{aligned}
        \Delta \bm x = [\Delta\bm p_{0}^\top, \Delta\bm v_{0}^\top, \Delta\bm q_{0}^\top, \Delta\bm \omega_{0}^\top, \Delta\bm d_{\bm p_0}^\top, \Delta\bm d_{\bm q_0}^\top, \Delta\bm d_{\bm p_1}^\top, \Delta\bm d_{\bm p_2}^\top]^\top
    \end{aligned}
\end{equation}
and the input offset around the equilibrium is expressed as
\begin{equation}
    \begin{aligned}
        \Delta\bm u = [(\Delta\bm u_1^{tot})\top, (\Delta\bm u_2^{tot})^\top]^\top
    \end{aligned}
\end{equation}
The model \eqref{finalmodel} is then linearized around the chosen equilibrium as
\begin{equation}\label{linearizedmodel}
    \begin{aligned}
    \Delta\dot{\bm p}_0 &= \Delta \bm v_0\\
    \Delta\dot{\bm v}_0 &= -\frac{1}{m_0} \sum_{i=1}^2{\Delta\bm u_i^{tot}} + \frac{1}{m_0} \sum_{i=0}^2{\Delta\bm{d}_{\bm{p}_i}}\\
    \Delta\dot{\bm q}_0 &= -\bm q_{0e}^\times \Delta\bm \omega_0\\
    \Delta\dot{\bm \omega}_0 &= \frac{1}{J_0} \sum_{i=1}^2{(-1)^{i} \rho_i \bm q_{0e}^\times (\Delta\bm u_i^{tot}-\Delta\bm d_{\bm p_i})} + \frac{1}{J_0} \bm q_{0e}^\times \Delta \bm d_{\bm q_0} + \frac{1}{J_0} \sum_{i=1}^2{(-1)^{i+1} \rho_i \bm \mu_{ie}^\times \Delta \bm q_0} - \frac{1}{J_0} \bm d_{\bm q_0, e}^\times \Delta \bm q_0\\ 
    \Delta\dot{\bm d}_{\bm p_0} &= \bm 0_{3 \times 1}\\
    \Delta\dot{\bm d}_{\bm q_0} &= \bm 0_{3 \times 1}\\
    \Delta\dot{\bm d}_{\bm p_1} &= \bm 0_{3 \times 1}\\
    \Delta\dot{\bm d}_{\bm p_2} &= \bm 0_{3 \times 1}\\
\end{aligned}
\end{equation}
The corresponding linearized observations are
\begin{equation}\label{linearizedobservations}
    \begin{aligned}
        \Delta \bm h_1 &= 
        \begin{bmatrix}
        \Delta \bm p_0 + \rho_1 \Delta \bm q_0 + \frac{l_1}{\|\bm \mu_{1e}\|} \bm q_{1e}^\times \bm q_{1e}^\times (\Delta \bm u_{1}^{tot} - \Delta \bm d_{\bm p_1})\\
          \Delta \bm p_0 - \rho_2 \Delta \bm q_0 + \frac{l_2}{\|\bm \mu_{2e}\|} \bm q_{2e}^\times \bm q_{2e}^\times (\Delta \bm u_{2}^{tot} - \Delta \bm d_{\bm p_2})\\
          \end{bmatrix}\\
        \Delta \bm h_2^r &= 
        \begin{bmatrix}
            \bm q_{0e}^\top \Delta \bm q_0\\
            \bm q_{0e}^\top \Delta \bm \omega_0\\
        \end{bmatrix}\\
        \Delta \bm h_3 &= 
        \begin{bmatrix}
            \bm q_{0e}^\top \Delta \bm d_{\bm q_0} + \bm d_{\bm q_0, e}^\top \Delta \bm q_0
        \end{bmatrix}
    \end{aligned}
    \end{equation}
    where $\bm q_{ie} = \bm \mu_{ie}/{\|\bm \mu_{ie}\|} \ (i=1,2)$ denotes the cable direction at the equilibrium.
    
Combining the linearized model \eqref{linearizedmodel} and the expressions of the cable force in \eqref{cableforceeq}, the system matrix $\bm A$ is computed as
\begin{equation}
    \begin{aligned}
        \bm A = 
        \begin{bmatrix}
            \bm A_{nom} & \bm A_{\bm d_{\bm p_0}} & \bm A_{\bm d_{\bm q_0}} & \bm A_{\bm d_{\bm p_1}} & \bm A_{\bm d_{\bm p_2}}\\
            \bm 0_{12\times 12} & \bm 0_{12\times 3} & \bm 0_{12\times 3} & \bm 0_{12\times 3} & \bm 0_{12\times 3}
        \end{bmatrix}
    \end{aligned}
\end{equation}
where $\bm A_{nom}$ represents the nominal system matrix.
$\bm A_{\bm d_{\bm p_0}}$, $\bm A_{\bm d_{\bm q_0}}$, $\bm A_{\bm d_{\bm p_1}}$, and $\bm A_{\bm d_{\bm p_2}}$ represent the input matrices of the disturbances $\Delta\bm d_{\bm p_0}$, $\Delta\bm d_{\bm q_0}$, $\Delta\bm d_{\bm p_1}$, and $\Delta\bm d_{\bm p_2}$, respectively. Here
\begin{equation}
    \begin{aligned}
    \bm A_{nom} &=
        \begin{bmatrix}
            \bm 0_{3\times 3} & \bm I_3 & \bm 0_{3\times 3} & \bm 0_{3\times 3}\\
            \bm 0_{3\times 3} & \bm 0_{3\times 3} & \bm 0_{3\times 3} & \bm 0_{3\times 3}\\
            \bm 0_{3\times 3} & \bm 0_{3\times 3} & \bm 0_{3\times 3} & -\bm q_{0e}^\times\\
            \bm 0_{3\times 3} & \bm 0_{3\times 3} & -\frac{n_0}{J_0} \bm q_{0e}^\times & \bm 0_{3\times 3}
        \end{bmatrix}\\
        \bm A_{\bm d_{\bm p_0}} &=
        \begin{bmatrix}
            \bm 0_{3\times 3} & \frac{1}{m_0}\bm I_3 & \bm 0_{3\times 3} & \bm 0_{3\times 3}
        \end{bmatrix}^\top\\
        \bm A_{\bm d_{\bm q_0}} &=
        \begin{bmatrix}
            \bm 0_{3\times 3}  & \bm 0_{3\times 3} & \bm 0_{3\times 3} & -\frac{1}{J_0} \bm q_{0e}^\times
        \end{bmatrix}^\top\\
        \bm A_{\bm d_{\bm p_1}} &=
        \begin{bmatrix}
            \bm 0_{3\times 3} & \frac{1}{m_0}\bm I_3 & \bm 0_{3\times 3} & -\frac{\rho_1}{J_0} \bm q_{0e}^\times
        \end{bmatrix}^\top\\
        \bm A_{\bm d_{\bm p_2}} &=
        \begin{bmatrix}
            \bm 0_{3\times 3} & \frac{1}{m_0}\bm I_3 & \bm 0_{3\times 3} & \frac{\rho_2}{J_0} \bm q_{0e}^\times
        \end{bmatrix}^\top\\
    \end{aligned}
\end{equation}
Based on the linearized observations \eqref{linearizedobservations}, the output matrix $\bm C$ is computed as
\begin{equation}
    \begin{aligned}
        \bm C = 
        \left[\begin{array}{ccccc}
            \bm C_{nom} & \bm C_{\bm d_{\bm p_0}} & \bm C_{\bm d_{\bm q_0}} & \bm C_{\bm d_{\bm p_1}} & \bm C_{\bm d_{\bm p_2}}\\
            \hdashline
            \bm C_{1}^{\bm h_3} & \bm 0_{1\times 3} & \bm C_{2}^{\bm h_3} & \bm 0_{1\times 3} & \bm 0_{1\times 3}
        \end{array}
        \right]
    \end{aligned}
\end{equation}
where $\bm C_{nom}$ represents the nominal output matrix.
$\bm C_{\bm d_{\bm p_0}}$, $\bm C_{\bm d_{\bm q_0}}$, $\bm C_{\bm d_{\bm p_1}}$, and $\bm C_{\bm d_{\bm p_2}}$ represent the partial derivatives of the outputs $\Delta \bm h_1$ and $\Delta \bm h_2^r$ with respect to the disturbances $\Delta\bm d_{\bm p_0}$, $\Delta\bm d_{\bm q_0}$, $\Delta\bm d_{\bm p_1}$, and $\Delta\bm d_{\bm p_2}$, respectively. 
$\bm C_{1}^{\bm h_3}$ and $\bm C_{2}^{\bm h_3}$ represent the partial derivatives of the output $\Delta \bm h_3$ with respect to the payload state and the disturbance $\Delta \bm d_{\bm q_0}$, respectively.
Here
\begin{equation}
    \begin{aligned}
    \bm C_{nom} &=
        \begin{bmatrix}
            \bm I_{3} & \bm 0_{3\times 3} & \rho_1 \bm I_{3} & \bm 0_{3\times 3}\\
            \bm I_{3} & \bm 0_{3\times 3} & -\rho_2 \bm I_{3} & \bm 0_{3\times 3}\\
            \bm 0_{1\times 3} & \bm 0_{1\times 3} & \bm q_{0e}^\top & \bm 0_{1 \times 3}\\
            \bm 0_{1\times 3} & \bm 0_{1\times 3} & \bm 0_{1\times 3} & \bm q_{0e}^\top
        \end{bmatrix}\\
    \bm C_{\bm d_{\bm p_0}} &=
        \bm 0_{8\times 3}\\
    \bm C_{\bm d_{\bm q_0}} &=
    \bm 0_{8\times 3}\\
     \bm C_{\bm d_{\bm p_1}} &=
    \begin{bmatrix}
        -\frac{l_1}{\|\bm \mu_{1e}\|}\bm q_{1e}^\times\bm q_{1e}^\times & \bm 0_{3\times 3} & \bm 0_{3\times 1} & \bm 0_{3\times 1}
    \end{bmatrix}^\top\\
    \bm C_{\bm d_{\bm p_2}} &=
    \begin{bmatrix}
        \bm 0_{3\times 3} & -\frac{l_2}{\|\bm \mu_{2e}\|}\bm q_{2e}^\times\bm q_{2e}^\times & \bm 0_{3\times 1} & \bm 0_{3\times 1}
    \end{bmatrix}^\top\\
    \bm C_{1}^{\bm h_3} &=
    \begin{bmatrix}
        \bm 0_{1\times 3} & \bm 0_{1\times 3} & \bm d_{\bm q_0,e}^\top & \bm 0_{1\times 3}
    \end{bmatrix}\\
    \bm C_{2}^{\bm h_3} &= \bm q_{0e}^\top 
    \end{aligned}
\end{equation}
Since the system matrix $\bm A$ and the output matrix $\bm C$ are assumed to be time invariant, the observability of the linearized system can be judged by simply utilizing the observability matrix.
Using the symbolic toolbox in Matlab, it can be found that
\begin{equation}
    \begin{aligned}
        \mathrm{rank}\left(
        \begin{bmatrix}
            \bm C\\
            \bm C \bm A\\
            \bm C \bm A^2
        \end{bmatrix}
        \right)
        =
        \mathrm{rank}\left(
        \begin{bmatrix}
            \bm C\\
            \bm C \bm A\\
            \bm C \bm A^2\\
            \bm C \bm A^3
        \end{bmatrix}
        \right)
        = 19
    \end{aligned}
\end{equation}
According to the observability rank criterion (ORC) in \cite{martinelli2020observability},
the rank of the observability matrix $\mathcal{O}$ can be verified to be $19$,
which is smaller than the system dimension $24$.
Therefore, the states of the linearized system \eqref{linearizedmodel} are not observable, i.e., the payload pose is not observable when all types of the disturbances $(\bm d_{\bm p_0}, \bm d_{\bm q_0}, \bm d_{\bm p_1}, \bm d_{\bm p_2})$ are exerted on the two-drone-bar system.

However, when one or part of the multi-source disturbances takes effect,
it is still uncertain whether the payload pose is observable or not.
To answer the above question,
we traverse all the possible disturbance combinations and test the observability by following the ORC in \cite{martinelli2020observability}.
The results are recorded in Table \ref{tab1}.
It can be concluded from Table \ref{tab1} that
the payload pose in the model \eqref{linearizedmodel} is observable under mild conditions when two or fewer types of lumped disturbances $(\bm d_{\bm p_0}, \bm d_{\bm q_0}, \bm d_{\bm p_1}, \bm d_{\bm p_2})$ are exerted on the whole system.
Based on the above conclusion, 
we can further infer that
if the disturbances on the payload, such as mass uncertainty and rotational inertia uncertainty,
are carefully calibrated offline,
it is possible to design filters to accurately estimate the payload pose when the wind disturbance on the payload is small.
Therefore, we follow this idea and develop a filter in the subsequent section.

\begin{table*}[!ht]
  \centering
  \caption{Observability analysis results}
  \label{tab1}
  \begin{tabular}{c c c c c c}  
    \hline
    Disturbance combination & System matrix & Output matrix & State dim & ORC & Observable \\ 
    \hline
    null & $\bm A_{nom}$ & $\bm C_{nom}$ & 12 & 12 & Yes \\ 
    \hline
    $\bm d_{\bm p_0}$ & $\begin{bmatrix}
        \bm A_{nom} & \bm A_{\bm d_{\bm p_0}}\\
        \bm 0_{3\times 12} & \bm 0_{3\times 3}
    \end{bmatrix}$ & $\begin{bmatrix}
        \bm C_{nom} & \bm C_{\bm d_{\bm p_0}}
    \end{bmatrix}$ & 15 & 15 & Yes \\ 
    \hline
    $\bm d_{\bm q_0}$ & $\begin{bmatrix}
        \bm A_{nom} & \bm A_{\bm d_{\bm q_0}}\\
        \bm 0_{3\times 12} & \bm 0_{3\times 3}
    \end{bmatrix}$ & 
    $\begin{bmatrix}
        \bm C_{nom} & \bm C_{\bm d_{\bm q_0}}\\
        \bm C_1^{\bm h_3} & \bm C_2^{\bm h_3}
    \end{bmatrix}$
    & 15 & 15 & Yes \\ 
    \hline
    $\bm d_{\bm p_i} \ (i=1 \ \text{or} \ 2)$ & $\begin{bmatrix}
        \bm A_{nom} & \bm A_{\bm d_{\bm p_i}}\\
        \bm 0_{3\times 12} & \bm 0_{3\times 3}
    \end{bmatrix}$ & 
    $\begin{bmatrix}
        \bm C_{nom} & \bm C_{\bm d_{\bm p_i}}\\
    \end{bmatrix}$
    & 15 & 15 & Yes \\ 
    \hline
    $\bm d_{\bm p_0}, \bm d_{\bm q_0}$ & $\begin{bmatrix}
        \bm A_{nom} & \bm A_{\bm d_{\bm p_0}} & \bm A_{\bm d_{\bm q_0}}\\
        \bm 0_{6\times 12} & \bm 0_{6\times 3} & \bm 0_{6\times 3}
    \end{bmatrix}$ & 
    $\begin{bmatrix}
        \bm C_{nom} & \bm C_{\bm d_{\bm p_0}} & \bm C_{\bm d_{\bm q_0}}\\
        \bm C_1^{\bm h_3} & \bm 0_{1\times 3} & \bm C_2^{\bm h_3}
    \end{bmatrix}$
    & 18 & 18 & Yes \\
    \hline
    $\bm d_{\bm p_0}, \bm d_{\bm p_i} \ (i=1 \ \text{or} \ 2)$ & $\begin{bmatrix}
        \bm A_{nom} & \bm A_{\bm d_{\bm p_0}} & \bm A_{\bm d_{\bm p_i}}\\
        \bm 0_{6\times 12} & \bm 0_{6\times 3} & \bm 0_{6\times 3}
    \end{bmatrix}$ & 
    $\begin{bmatrix}
        \bm C_{nom} & \bm C_{\bm d_{\bm p_0}} & \bm C_{\bm d_{\bm p_i}}
    \end{bmatrix}$
    & 18 & 18 & Yes \\ 
    \hline
    $\bm d_{\bm q_0}, \bm d_{\bm p_i} \ (i=1 \ \text{or} \ 2)$ & $\begin{bmatrix}
        \bm A_{nom} & \bm A_{\bm d_{\bm q_0}} & \bm A_{\bm d_{\bm p_i}}\\
        \bm 0_{6\times 12} & \bm 0_{6\times 3} & \bm 0_{6\times 3}
    \end{bmatrix}$ & 
    $\begin{bmatrix}
        \bm C_{nom} & \bm C_{\bm d_{\bm q_0}} & \bm C_{\bm d_{\bm p_i}}\\
        \bm C_1^{\bm h_3} & \bm C_2^{\bm h_3} & \bm 0_{1\times 3}
    \end{bmatrix}$
    & 18 & 18 & Yes \\ 
    \hline
    $\bm d_{\bm p_1}, \bm d_{\bm p_2}$ & $\begin{bmatrix}
        \bm A_{nom} & \bm A_{\bm d_{\bm p_1}} & \bm A_{\bm d_{\bm p_2}}\\
        \bm 0_{6\times 12} & \bm 0_{6\times 3} & \bm 0_{6\times 3}
    \end{bmatrix}$ & 
    $\begin{bmatrix}
        \bm C_{nom} & \bm C_{\bm d_{\bm p_1}} & \bm C_{\bm d_{\bm p_2}}\\
    \end{bmatrix}$
    & 18 & 18 & Yes \\ 
    \hline
    $\bm d_{\bm p_0}, \bm d_{\bm q_0}, \bm d_{\bm p_i} \ (i=1 \ \text{or} \ 2)$ & $\begin{bmatrix}
        \bm A_{nom} & \bm A_{\bm d_{\bm p_0}} & \bm A_{\bm d_{\bm q_0}} & \bm A_{\bm d_{\bm p_i}}\\
        \bm 0_{9\times 12} & \bm 0_{9\times 3} & \bm 0_{9\times 3} & \bm 0_{9\times 3}
    \end{bmatrix}$ & 
    $\begin{bmatrix}
        \bm C_{nom} & \bm C_{\bm d_{\bm p_0}} & \bm C_{\bm d_{\bm q_0}} & \bm C_{\bm d_{\bm p_i}}\\
        \bm C_1^{\bm h_3} & \bm 0_{1\times 3} & \bm C_2^{\bm h_3} & \bm 0_{1\times 3}
    \end{bmatrix}$
    & 21 & 19 & No \\ 
    \hline
    $\bm d_{\bm p_0}, \bm d_{\bm p_1}, \bm d_{\bm p_2}$ & $\begin{bmatrix}
        \bm A_{nom} & \bm A_{\bm d_{\bm p_0}} & \bm A_{\bm d_{\bm p_1}} & \bm A_{\bm d_{\bm p_2}}\\
        \bm 0_{9\times 12} & \bm 0_{9\times 3} & \bm 0_{9\times 3} & \bm 0_{9\times 3}
    \end{bmatrix}$ & 
    $\begin{bmatrix}
        \bm C_{nom} & \bm C_{\bm d_{\bm p_0}} & \bm C_{\bm d_{\bm p_1}} & \bm C_{\bm d_{\bm p_2}}
    \end{bmatrix}$
    & 21 & 18 & No \\ 
    \hline
    $\bm d_{\bm q_0}, \bm d_{\bm p_1}, \bm d_{\bm p_2}$ & $\begin{bmatrix}
        \bm A_{nom} & \bm A_{\bm d_{\bm q_0}} & \bm A_{\bm d_{\bm p_1}} & \bm A_{\bm d_{\bm p_2}}\\
        \bm 0_{9\times 12} & \bm 0_{9\times 3} & \bm 0_{9\times 3} & \bm 0_{9\times 3}
    \end{bmatrix}$ & 
    $\begin{bmatrix}
        \bm C_{nom} & \bm C_{\bm d_{\bm q_0}} & \bm C_{\bm d_{\bm p_1}} & \bm C_{\bm d_{\bm p_2}}\\
        \bm C_1^{\bm h_3} & \bm C_2^{\bm h_3} & \bm 0_{1\times 3} & \bm 0_{1\times 3}
    \end{bmatrix}$
    & 21 & 19 & No \\ 
    \hline
  \end{tabular}
\end{table*}

\begin{rem}
    We choose to focus on the disturbances on the drones because the thrust uncertainty and inter-vehicle aerodynamic effect in the disturbances $\bm d_{\bm p_i} \ (i=1,2)$ are difficult to model and calibrate offline.
\end{rem}

\section{Disturbance Observer-based Error-state Extended Kalman Filter Design}
In this section, fundamental knowledge about the ESEKF is reviewed.
Then, the DO-based ESEKF is designed for the system.

\subsection{ESEKF Fundamentals}\label{framework}
Let $\mathcal M$ be an $n$-dimensional manifold  with a boxplus operator $\boxplus: \mathcal M \times \mathbb R^n \to \mathcal{M}$ such that, for every
 $\bm x \in \mathcal M$,
 $\bm x \boxplus \bm \delta$ is a parametrization
 of a neighborhood of $\bm x$ for a sufficiently small $\bm \delta \in \mathsf{T}_{\bm x}\mathcal{M}\cong\mathbb R^n$.
 We also define the inverse operator $\boxminus: \mathcal{M} \times \mathcal{M} \to \mathbb R^n$ so
 that $\bm x \boxplus (\bm y \boxminus \bm x) = \bm y$ for sufficiently close $\bm x, \bm y \in \mathcal M$.
 This allows us to transcribe many algorithms from $\mathbb R^n$ to $\mathcal M$
 simply by replacing the vector space operators $+$ and $-$ with $\boxplus$ and $\boxminus$ (see \cite{hertzberg2013integrating} for more details).

Firstly, we consider the problem of estimating the state $\bm x$ of a system defined on an $n$-dimensional manifold $\mathcal{M}_s$  given its continuous dynamic equation $\bm f_c(\cdot)$ and the external observations $\bm z$. The system is described by the equations
\begin{equation}\label{continuousmodel}
    \begin{aligned}
        \dot{\bm x} &= \bm f_c(\bm x, \bm u)\\
        \bm z &= \bm h(\bm x)
    \end{aligned}
\end{equation}
For simplicity, we assume that $\bm z$ evolves in the Euclidean space,
which is consistent with the state estimation problem in this research.
Under the Gaussian random noise assumption,
the continuous model \eqref{continuousmodel} is then discretized at each measurement step $k$. 
Denote $\Delta t_k$ as the time interval from $k-1$ to $k$, and the discretized model is derived by assuming the first-order time derivative of the state is constant during one sampling period
\begin{equation}\label{ESEKFMODEL}
    \begin{aligned}
        \bm x_k &= \bm x_{k-1} \oplus [\Delta t_k \bm f_d(\bm x_{k-1}, \bm u_{k-1}, \bm w_{k-1})]\\
        \bm z_k &= \bm h(\bm x_k) + \bm n_k
    \end{aligned}
\end{equation}
 where the operation $\oplus$ denotes the "addition" of the state $\bm x_{k-1}$ and the perturbation vector,
 which naturally yields the state $\bm x_{k}$ remaining on the manifold $\mathcal M$.
 Based on this operation,
 the discrete dynamic equation $\bm f_d(\cdot)$ is defined.
 $\bm w_{k-1}$ is the discrete noise at the step $k-1$ with a Gaussian
 distribution $\mathcal{N}(\bm 0,\bm Q_{k-1})$, 
 and $\bm n_k$ is the discrete noise at the step $k$ with a Gaussian
 distribution $\mathcal{N}(\bm 0,\bm R_{k})$.

Next, we describe the estimation process of the error-state
 EKF (ESEKF).
 Denote the posterior state estimate at the step $k-1$ by $\bm x^+_{k-1}$ and its covariance by $\bm P_{k-1}^+$. 
 The propagation step propagates the state estimate to the step $k$ by
\begin{equation}\label{propagation}
    \begin{aligned}
 \bm x^-_{k} &= \bm x_{k-1}^+ \oplus [\Delta t_k \bm f_d(\bm x_{k-1}^+, \bm u_{k-1}, \bm 0)]\\
 \bm P^-_{k} &= \bm F_{\bm x_{k-1}} \bm P^+_{k-1}\bm F^\top_{\bm x_{k-1}} + \bm F_{\bm w_{k-1}} \bm Q_{k-1} \bm F^\top_{\bm w_{k-1}}
    \end{aligned}
\end{equation}
According to the results in \cite{he2023symbolic},
the propagation Jacobians in \eqref{propagation} are computed as
\begin{equation}
    \begin{aligned}
        \bm F_{\bm x_{k-1}} &= \bm G_{\bm x_{k-1}} + \Delta t \bm G_{\bm f_{k-1}} \left. \frac{\partial \bm f_d(\bm x_{k-1}^+\boxplus\delta \bm x, \bm u_{k-1}, \bm 0)}{\partial \delta \bm x}\right|_{\delta \bm x = \bm 0}\\
        \bm F_{\bm w_{k-1}} &= \Delta t \bm G_{\bm f_{k-1}} \left.\frac{\partial \bm f_d(\bm x_{k-1}^+, \bm u_{k-1}, \bm w)}{\partial \bm w}\right|_{\bm w=\bm 0}
    \end{aligned}
\end{equation}
where $\bm G_{\bm x_{k-1}}$ and $\bm G_{\bm f_{k-1}}$ are manifold-specific parts and can be calculated independent from the system dynamics $\bm f_d(\cdot)$.

With the measurement $\bm z_k$, the residual is calculated based on the observation function in \eqref{ESEKFMODEL} and linearized  as
\begin{equation}\label{posteriodistribution}
    \begin{aligned}
        \bm r_k &= \bm z_k - \bm h(\bm x^-_k)\\
        &\approx \bm H_k \delta \bm x_k + \bm n_k
    \end{aligned}
\end{equation}
where $\delta \bm x_k = \bm x_k \boxminus \bm x_k^-$ denotes the error state between the propagated state $\bm x_k^-$ and the groundtruth $\bm x_k$, and
\begin{equation}
    \begin{aligned}
        \bm H_k &= \left.\frac{\partial \bm h(\bm x_k^-\boxplus\delta \bm x)}{\partial \delta \bm x}\right|_{\delta \bm x = \bm 0}\\
    \end{aligned}
\end{equation}
As the prior distribution, the error state $\delta \bm x_k = \bm x_k \boxminus \bm x_k^-$ satisfies $\delta \bm x_k \sim \mathcal{N}(\bm 0, \bm P_k^-)$. 
The observation model in \eqref{posteriodistribution} gives the posterior distribution
\begin{equation}
    \begin{aligned}
        \bm n_k \approx \bm r_k - \bm H_k \delta \bm x_k \sim \mathcal{N}(\bm 0, \bm R_k)
    \end{aligned}
\end{equation}
Combining the prior and posterior distributions and following the Bayesian rule yields the following maximum a posterior (MAP) problem:
\begin{equation}\label{mapproblem}
    \begin{aligned}
    \underset{\delta \bm x_k}{\mathrm{argmin}} \left(\|\delta \bm x_k\|_{\bm P_k^-}^2 + \| \bm r_k-\bm H_k \delta \bm x_k\|_{\bm R_k}^2 
    \right)
    \end{aligned}
\end{equation}
where $\|\bm x\|_{\bm A}^2=\bm x^\top \bm A^{-1} \bm x$.
The optimization problem in \eqref{mapproblem} is a standard quadratic programming and the optimal solution $\delta \bm x_k^o$ corresponds to the Kalman update:
\begin{equation}\label{errorstateupdate}
    \begin{aligned}
        \delta \bm x_k^o &= \bm K_k (\bm z_k - \bm h(\bm x^-_k))\\
        \bm K_k &= \bm P_k^-\bm H_k^\top (\bm H_k \bm P_k^-\bm H_k^\top + \bm R_k)^{-1}\\
    \end{aligned}
\end{equation}
Finally, the state is updated as $\bm x_k^+ = \bm x_k^-\boxplus \delta \bm x_k^o$,
and the covariance is updated as
\begin{equation}
    \begin{aligned}
        \bm P_k &= (\bm I - \bm K_{k} \bm H_k) \bm P_k^-\\
        \bm P_k^+ &= \bm J_k \bm P_k \bm J_k^\top\\
    \end{aligned}
\end{equation}
where $\bm J_k$ is the projection matrix and can also be calculated independently.
\begin{equation}
    \begin{aligned}
        \bm J_k = \left.\frac{\partial((\bm x_k^-\boxplus \delta \bm x)\boxminus \bm x_k^+)}{\partial \delta \bm x}\right|_{\delta \bm x=\delta \bm x_k^o}
    \end{aligned}
\end{equation}

\subsection{DO-based ESEKF for the system}
We now detail the proposed filter for the two-drone-bar system based on the framework in the subsection \ref{framework}.
With the multi-source disturbances exerted only on the drones,
the dynamic model \eqref{fulldynamics} affected by Gaussian distribution noise is derived as

\begin{equation}\label{noisedynamics}
\begin{aligned}
		&m_T \dot{\bm{v}}_0 + \sum_{i=1}^{2} m_i [(-1)^i \rho_i (\bm q_0^{\times} \dot{\bm{\omega}}_0 + \|\bm{\omega}_0\|^2  \bm q_0)+ l_i \bm{q}_i^{\times} \dot{\bm{\omega}}_i + l_i \|\bm{\omega}_i\|^2  \bm q_i]
        = m_T g \bm{e}_3 + \sum_{i=1}^{2} (\bm u_i + \bm d_{\bm p_i}) + \bm w_{\bm p_0},\\
		&\bar{J}_0 \dot{\bm{\omega}}_0 + \sum_{i=1}^{2} (-1)^{i+1} m_i \rho_i \bm q_0^\times (\dot{\bm{v}}_0 + l_i \bm{q}_i^{\times} \dot{\bm{\omega}}_i + l_i \| \bm{\omega}_i \|^2 \bm{q}_i) = \sum_{i=1}^{2} (-1)^{i+1} \rho_i \bm q_0^\times (\bm u_i + \bm d_{\bm p_i} + m_i g \bm{e}_3) + \bm q_0^\times \bm w_{\bm q_0},\\
		&m_i l_i \dot{\bm{\omega}}_i - m_i \bm{q}_i^{\times} [\dot{\bm{v}}_0 + (-1)^{i} \rho_i \bm q_0^{\times} \dot{\bm{\omega}}_0 - \rho_i \|\bm \omega_0\|^2 \bm{q}_0] = - \bm{q}_i^{\times} (\bm u_i + \bm d_{\bm p_i} + m_i g \bm{e}_3), \quad i = 1, 2
	\end{aligned}
\end{equation}
where 
$\bm w_{\bm p_0}$ and $\bm w_{\bm q_0}$ denote the process noise on the payload translational channel and rotational channel, respectively.
The dynamic model \eqref{noisedynamics} can be reorganized into the compact form 
\begin{equation}\label{compact_form}
\bm M \left[ \dot{\bm v}_0^\top, \dot{\bm \omega}_0^\top, \dot{\bm \omega}_1^\top, \dot{\bm \omega}_2^\top\right]^\top = \bm F
\end{equation}
Readers may refer to Appendix C for the detailed formulation of $\bm M$ and $\bm F$. 

In the spirit of disturbance observer (DO), 
the disturbances on the drones $\bm d_{\bm p_i} \ (i=1,2)$ are modelled as random-walk biases driven by $\bm w_{\bm d_{\bm p_i}} \ (i=1,2)$,
which are selected based on the disturbance magnitude.
Then, the full state for estimation is defined as
\begin{equation}\label{x}
    \begin{aligned}
        \bm x = 
        \left[
            \bm p_0^\top, \bm v_0^\top, \bm q_0^\top, \bm \omega_0^\top, \bm q_1^\top, \bm \omega_1^\top, \bm q_2^\top, \bm \omega_2^\top, \bm d_{\bm p_1}^\top, \bm d_{\bm p_2}^\top
        \right]^\top
    \end{aligned}
\end{equation}
which evolves on the Cartesian product manifold
$$ (\mathbb R^3)^2 \times (\mathsf{TS}^2)^3 \times (\mathbb R^3)^2.$$
The $\boxplus/\boxminus$ and $\oplus$ operations and their partial differentiation on the manifolds $\mathbb R^n$ and $\mathsf{TS}^2$ are summarized in Appendix A.

To apply the framework in the subsection \ref{framework},
we need to compute the manifold-specific part and system-specific part separately.

 For the manifold-specific part,
 the relevant matrices are computed as
 \begin{equation}
     \begin{aligned}
         \bm G_{\bm x_\tau} &= \mathrm{blkdiag}(\bm G_{\bm x_\tau}^{\bm p_0, \bm v_0}, \bm G_{\bm x_\tau}^{\bm q_0, \bm \omega_0}, \bm G_{\bm x_\tau}^{\bm q_1, \bm \omega_1}, \bm G_{\bm x_\tau}^{\bm q_2, \bm \omega_2}, \bm G_{\bm x_\tau}^{\bm d_{\bm p_1}, \bm d_{\bm p_2}})\\
         \bm G_{\bm f_\tau} &= \mathrm{blkdiag}(\bm G_{\bm f_\tau}^{\bm p_0, \bm v_0}, \bm G_{\bm f_\tau}^{\bm q_0, \bm \omega_0}, \bm G_{\bm f_\tau}^{\bm q_1, \bm \omega_1}, \bm G_{\bm f_\tau}^{\bm q_2, \bm \omega_2}, \bm G_{\bm f_\tau}^{\bm d_{\bm p_1}, \bm d_{\bm p_2}})\\
         \bm J_{\tau} &=
         \mathrm{blkdiag}(
         \bm J_\tau^{\bm p_0, \bm v_0},
         \bm J_\tau^{\bm q_0, \bm \omega_0},
         \bm J_\tau^{\bm q_1, \bm \omega_1},
         \bm J_\tau^{\bm q_2, \bm \omega_2},
         \bm J_\tau^{\bm d_{\bm p_1}, \bm d_{\bm p_2}})
     \end{aligned}
 \end{equation}
 where $\tau$ denotes the time step of the measurement and the detailed expressions of $\bm G_{(\cdot)}^{(\cdot)}$ and $\bm J_{(\cdot)}^{(\cdot)}$ are presented in Appendix B.
 
 For the system-specific part, by combining the system kinematics \eqref{kinematics} and dynamics \eqref{noisedynamics},
the state model is cast into the following canonical form:
\begin{equation}\label{canonical}
    \begin{aligned}
        &\bm f_d(\bm x, \bm u, \bm w) = 
        \Big[
            \bm v_0^\top,
            (\bm M^{-1}\bm F)_{1:3}^\top,
            \bm \omega_0^\top,
            (\bm M^{-1}\bm F)_{4:6}^\top, \bm \omega_1^\top,
            (\bm M^{-1}\bm F)_{7:9}^\top,
            \bm \omega_2^\top,
            (\bm M^{-1}\bm F)_{10:12}^\top,
            \bm w_{\bm d_{\bm p_1}}^\top,
            \bm w_{\bm d_{\bm p_2}}^\top
        \Big]^\top
    \end{aligned}
\end{equation}
where $(\bm M^{-1}\bm F)_{i:j}$ denotes the vector constituting by the $i$th to the $j$th elements of the vector $\bm M^{-1} \bm F$.
The control input is denoted as
\begin{equation}\label{control_input}
\bm u = [\bm u_1^\top, \bm u_2^\top]^\top
\end{equation}
 and the process noise is denoted as 
    \begin{equation}\label{process_noise}
        \bm w = [\bm w_{\bm p_0}^\top, \bm w_{\bm q_0}^\top, \bm w_{\bm d_{\bm p_1}}^\top, \bm w_{\bm d_{\bm p_2}}^\top]^\top.
    \end{equation}
In addition, by using the error state \eqref{errorstateupdate} in the filter,
we can avoid considering the manifold constraints in the observation \eqref{measurements}.
Under experimental conditions, we choose the positions $\bm p_i \ (i=1,2)$ and velocities $\bm v_i \ (i=1,2)$ of the two drones as the system outputs, i.e.,
\begin{equation}\label{actual_measurements}
    \begin{aligned}
        \bm h(\bm x) = 
        \left[
            \bm p_1^\top, \bm p_2^\top, \bm v_1^\top, \bm v_2^\top
        \right]^\top
    \end{aligned}
\end{equation}
and the corresponding measurement noise is denoted as 
$\bm n = \left[\bm n_{\bm p_1}^\top, \bm n_{\bm p_2}^\top, \bm n_{\bm v_1}^\top, \bm n_{\bm v_2}^\top\right]^\top$.
Accordingly, the system-specific partial differentiations,
 including 
 $\left.\frac{\partial \bm f_d(\bm x \boxplus \delta \bm x, \bm u, \bm 0)}{\partial \delta \bm x}\right|_{\delta \bm x = \bm 0}$,
 $\left.\frac{\partial\bm f_d(\bm x, \bm u, \bm w)}{\partial \bm w}\right|_{\bm w = \bm 0}$,
 and $\left.\frac{\partial \bm h(\bm x \boxplus \delta \bm x)}{\partial \delta \bm x}\right|_{\delta \bm x = \bm 0}$
 can be calculated,
 whose expressions are presented in Appendix D.

\begin{figure}
    \centering
    \includegraphics[width=3.5in]{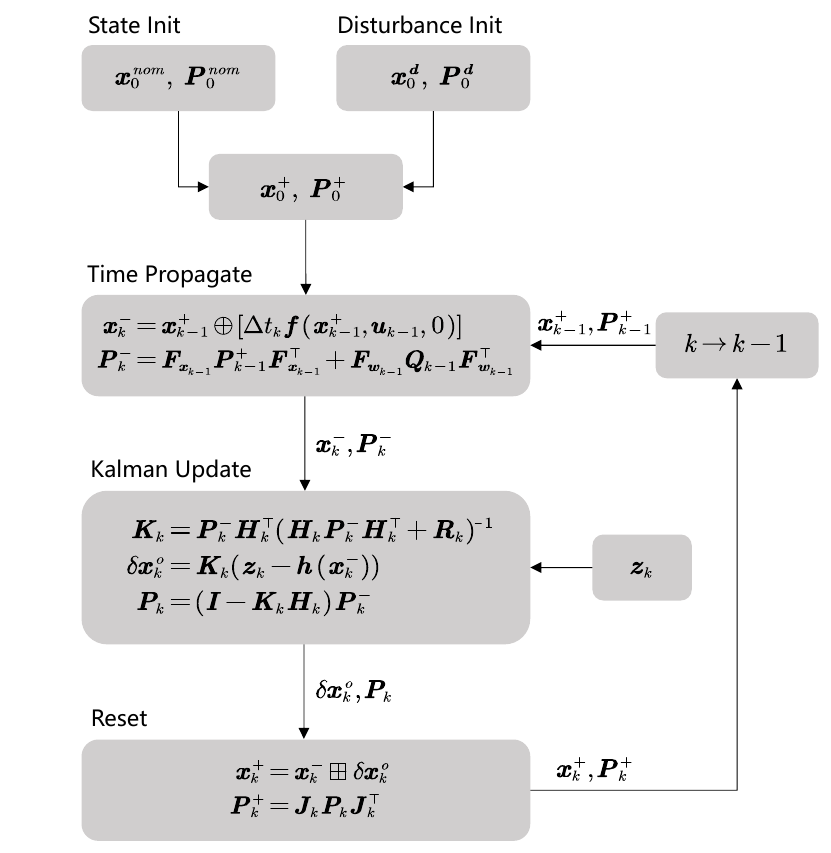}
    \caption{DO-based ESEKF workflow}
    \label{滤波器流程}
\end{figure}

 Based on the canonical representation of the system \eqref{canonical} 
 and the respective partial differentiations,
 the DO-based ESEKF can be completely established and the filter workflow is shown in Fig \ref{滤波器流程},
 where $(\cdot)^{nom}_0$ denotes the initial value related to the system state and $(\cdot)^{\bm d}_0$ denotes the initial value related to the disturbance.
 Although the filter is designed for the case where only the disturbances on the drones are considered, 
 similar procedures can also be applied to other observable scenarios in Table \ref{tab1} for accurate state and disturbance estimation.

\begin{rem}
    We do not employ the model \eqref{finalmodel} and the observation \eqref{finaloutput} for state estimation due to the correlation between the process noise and the measurement noise. 
    This correlation violates the independence assumption of the noise in the basic model \eqref{ESEKFMODEL}, which might introduce bias and degrade estimation performance.
\end{rem}

\begin{rem}
   The significance of our proposed DO-based ESEKF method lies in that it enables accurate estimation of the payload pose while relying solely on the odometry information of the drones, providing theoretical guidance for uncertainty calibration in practical engineering and laying a solid foundation for sensorless payload pose manipulation tasks in the future.
\end{rem}

\section{Simulation Study}\label{simulation}
In this section, we test our developed DO-based ESEKF in numerical simulations.
System and estimation parameters are listed in Table \ref{tab2}, 
where the matrix $\bm R_x(\theta)$ in the table is constructed as
\begin{equation}
    \bm R_x (\theta)
= 
\begin{bmatrix}
    1 & 0 & 0\\
    0 & \cos\theta & -\sin\theta\\
    0 & \sin\theta & \cos\theta
\end{bmatrix}
\end{equation}

\begin{table}[!htbp]
	\centering
	\caption{Parameters used in the simulations}
	\resizebox{!}{!}{
		\begin{tabular}{c  c | c  c} 
			\hline 
			\hline
			\multicolumn{4}{c}{System Parameters}\\
			\hline
                \multicolumn{2}{c}{Payload} & \multicolumn{2}{c}{Drones}\\
                \hline
                $m_0$ & $0.445, kg$ & $m_1$ & $0.900, kg$\\
			\hline
                $J_0$ & $0.148, kg\cdot m^2$ & $m_2$ & $0.900, kg$\\
                \hline
                $\rho_1$ & $1, \mathrm{m}$ & $l_1$ & $1, m$\\
                \hline 
                $\rho_2$ & $1, \mathrm{m}$ & $l_2$ & $1, m$\\
                \hline
			\multicolumn{4}{c}{Estimation Parameters}\\
			\hline
			\multicolumn{2}{c}{$\bm Q_k$} & \multicolumn{2}{c}{$1 \times {10}^{-1}\bm I_{12}$}\\
                \multicolumn{2}{c}{$\bm R_k$} & \multicolumn{2}{c}{$1 \times 10^{-2} \bm I_{12}$}\\
			\hline
			\multicolumn{4}{c}{Initial Conditions}\\
			\hline
			\multicolumn{2}{c}{$\bm p_0 (0)$} & \multicolumn{2}{c}{$[0.03, -0.05, -0.03]^\top, m$}\\
			\multicolumn{2}{c}{$\bm q_0 (0)$} & \multicolumn{2}{c}{$[1, 0, 0]^\top$}\\
            \multicolumn{2}{c}{$\bm q_1 (0)$} & \multicolumn{2}{c}{$\bm R_x(\pi/18)[0, 0, 1]^\top$}\\
            \multicolumn{2}{c}{$\bm q_2 (0)$} & \multicolumn{2}{c}{$\bm R_x(\pi/18)[0, 0, 1]^\top$}\\
            \multicolumn{2}{c}{$\bm v_0 (0)$} & \multicolumn{2}{c}{$[0, 0, 0]^\top, m/s$}\\
            \multicolumn{2}{c}{$\bm \omega_i (0) \ (i=0,1,2)$} & \multicolumn{2}{c}{$[0, 0, 0]^\top, rad/s$}\\
			\hline
			\hline
	\end{tabular}}\label{tab2}
\end{table}
The initial positions of the two drones can be computed according to the settings in Table \ref{tab2} as
\begin{equation}\label{initialpositions}
    \begin{aligned}
        \bm p_1(0) &= [1.030, 0.124, -0.955]^\top, m\\
        \bm p_2(0) &= [-0.970, 0.124, -0.955]^\top, m
    \end{aligned}
\end{equation}
The two drones in the simulations are subject to the disturbances as
\begin{equation}
    \begin{aligned}
        \bm d_{\bm p_1} &= [1, -1, 4]^\top, N\\
        \bm d_{\bm p_2} &= [1, 1, 2]^\top, N\\
    \end{aligned}
\end{equation}
To verify the effectiveness of the proposed DO-based ESEKF,
we equip the drones with the common PID controllers and simulate three different scenarios:
point stabilization, trajectory tracking, and payload disturbance injection.
In addition, white Gaussian noises are added to the drones' positions and velocities with the mean of zero and the variance of $0.01$,
producing the simulated measurements.
\subsection{Point Stabilization}
\begin{figure}[!ht]
    \centering
    \includegraphics[width=3.5in]{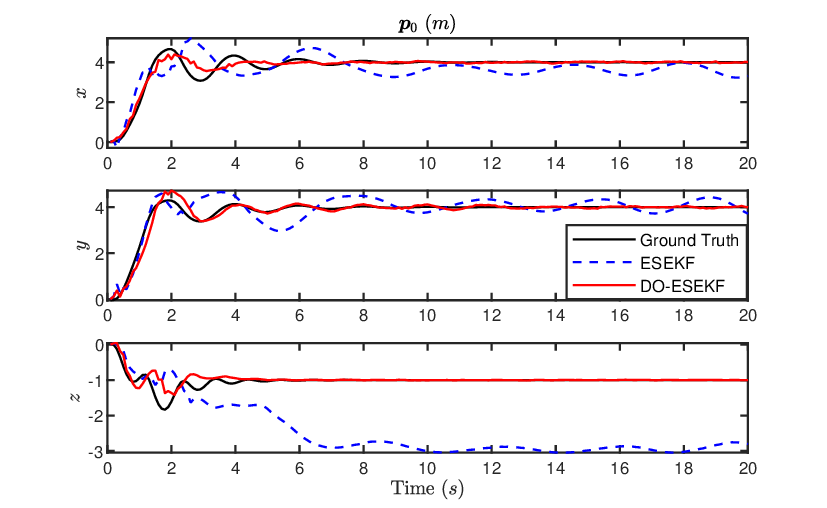}
    \caption{The estimation results of payload position in the simulation of point stabilization}
    \label{simulation_1_p0}
\end{figure}

\begin{figure}[!ht]
    \centering
    \includegraphics[width=3.5in]{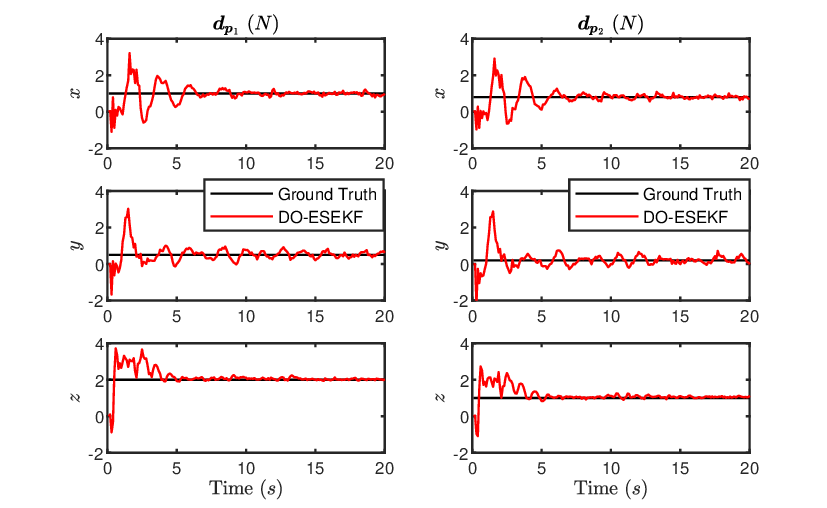}
    \caption{The estimation results of disturbances in the simulation of point stabilization}
    \label{simulation_1_dp}
\end{figure}

\begin{figure}[!ht]
    \centering
    \includegraphics[width=3.5in]{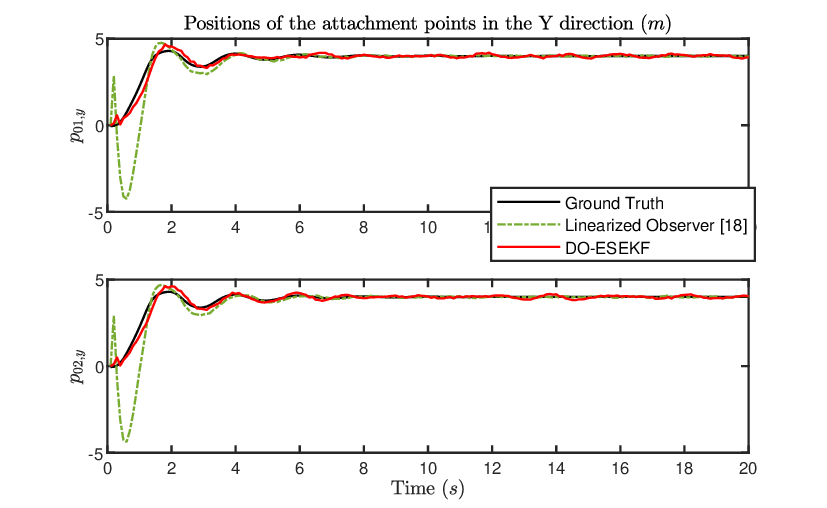}
    \caption{The comparison of payload estimation with the linearized method \cite{xu2024oscillation} in the simulation of point stabilization}
    \label{simulation_1_p01}
\end{figure}

In this scenario, the two drones are required to move to the desired positions
\begin{equation}\label{fixedpositions}
    \begin{aligned}
        \bm p_1^{des} &= [5, 4, -2]^\top, m\\
        \bm p_2^{des} &= [3, 4, -2]^\top, m\\
    \end{aligned}
\end{equation}
from the initial positions \eqref{initialpositions}.
The CoM position of the
 payload and the disturbances on the drones in the simulation are shown in Fig. \ref{simulation_1_p0} and Fig. \ref{simulation_1_dp}, respectively.
 We compare the state estimation results of the normal ESEKF and the proposed DO-based ESEKF in Fig. \ref{simulation_1_p0}, where the proposed DO-based ESEKF (red line) can converge quickly and closely track the ground truth (black line), showing excellent estimation accuracy. 
 Furthermore, the disturbances $\bm d_{\bm p_i} \ (i=1,2)$ can be accurately and quickly estimated as shown in Fig. \ref{simulation_1_dp}, which verifies the effectiveness of the proposed DO-based ESEKF.
 In contrast, the normal ESEKF (blue line) without disturbance estimation renders significant and persistent deviations from the ground truth in Fig. \ref{simulation_1_p0}, especially in the $Z$ direction, because it does not explicitly estimate and compensate for the disturbances on the UAVs.
 In addition, we test the method in \cite{xu2024oscillation} for more comparison.
 Since the linearized observer in \cite{xu2024oscillation} only estimates the positions of the attachment points on the payload in the Y direction,
 we utilize them for comparison with the proposed method. 
 As shown in Fig. \ref{simulation_1_p01},
 the DO-ESEKF exhibits lower estimation error throughout the stabilization process, especially during the initial phase.
 This demonstrates the superiority of the proposed method for state estimation on the nonlinear manifold, overcoming the limitation of local effectiveness of the linearized method.
 

\subsection{Trajectory Tracking}
\begin{figure}[!ht]
    \centering
    \includegraphics[width=3.5in]{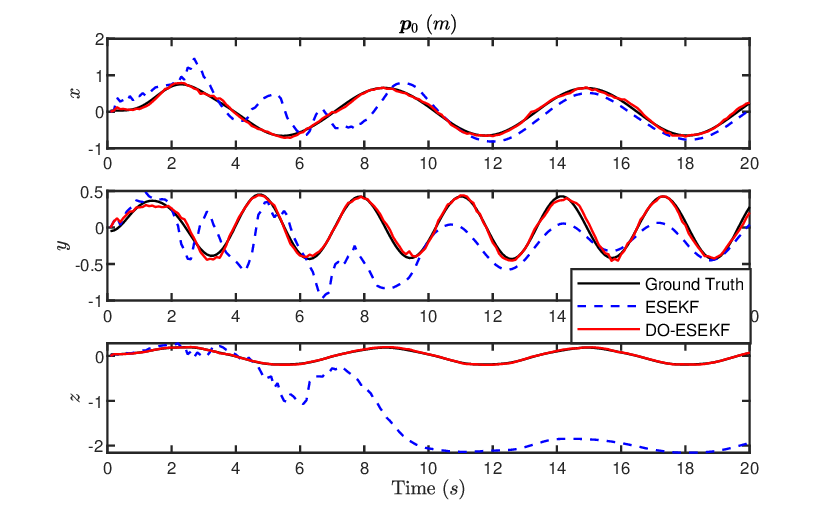}
    \caption{The estimation results of payload position in the simulation of trajectory tracking}
    \label{simulation_2_p0}
\end{figure}

\begin{figure}[!ht]
    \centering
    \includegraphics[width=3.5in]{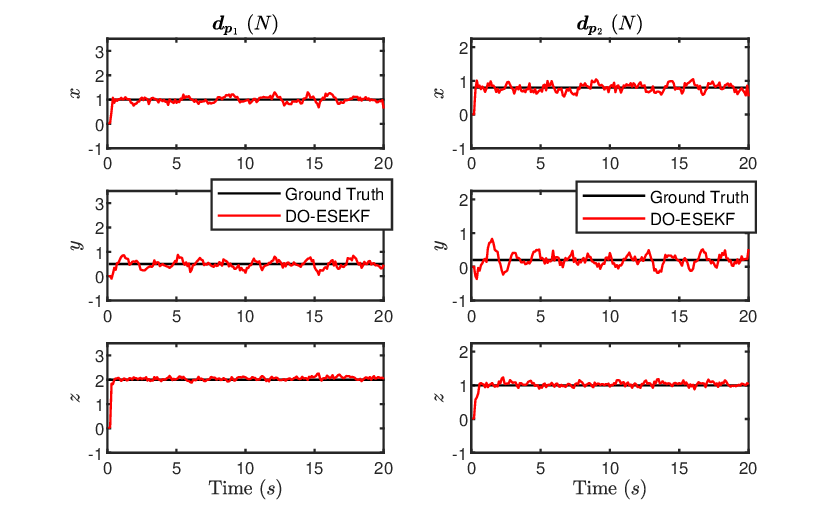}
    \caption{The estimation results of disturbances in the simulation of trajectory tracking}
    \label{simulation_2_dp}
\end{figure}

In this scenario, the two drones are required to fly along the agile trajectories  similar to the  figure-eight maneuver, 
which are set as
\begin{equation}
    \begin{aligned}
        \bm p_1^{des} (t) &= [1+0.7\sin t, 0.7 \sin t \cos t, -1+0.2\sin t]^\top, m\\
        \bm p_2^{des} (t) &= [-1+0.7\sin t, 0.7 \sin t \cos t, -1+0.25\sin t]^\top, m\\
    \end{aligned}
\end{equation}
Fig. \ref{simulation_2_p0} and Fig. \ref{simulation_2_dp} show the CoM position of the payload and the disturbances $\bm d_{\bm p_i} \ (i=1,2)$ during the aggressive trajectory tracking. 
It can be seen from Fig. \ref{simulation_2_p0} that DO-ESEKF (red line) maintains close alignment with the ground truth (black line) even under high-dynamic maneuvers while the standard ESEKF (blue line) shows notable tracking deviations due to unaddressed disturbances.
The disturbance estimates (red line) in 
Fig. \ref{simulation_2_dp} converge rapidly to the true values (black line) during the trajectory tracking, confirming the efficiency of the proposed method in transient conditions.

\begin{figure}[!ht]
    \centering
    \includegraphics[width=3.5in]{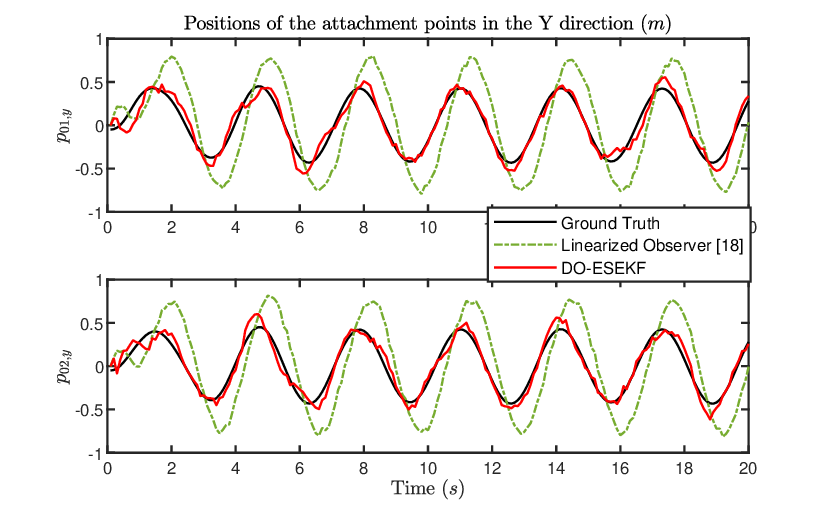}
    \caption{The comparison of payload estimation with the linearized method \cite{xu2024oscillation} in the simulation of trajectory tracking}
    \label{simulation_2_p01}
\end{figure}

The additional comparison with the method in \cite{xu2024oscillation} is shown in Fig. \ref{simulation_2_p01}, where DO-ESEKF yields significantly lower errors, particularly in high-curvature trajectory segments, illustrating the necessity of nonlinear filtering for agile motions.


\subsection{Payload Disturbance Injection}
\begin{figure}[!h]
    \centering
    \includegraphics[width=3.5in]{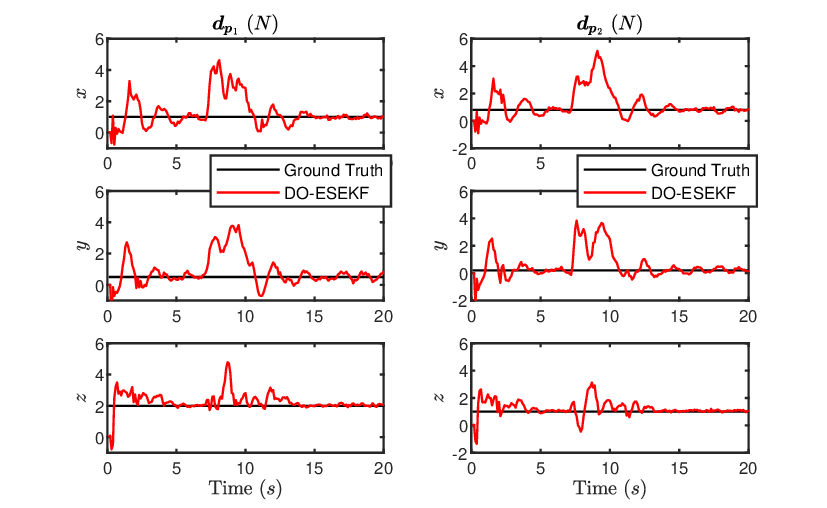}
    \caption{The estimation results of disturbances in the simulation of payload disturbance injection}
    \label{simulation_3_dp}
\end{figure}

Different from the above two scenarios,
we introduce the disturbances on the payload to test the robustness of the proposed method against model simplifications (wind disturbances $\bm d_{\bm p_0}$ and $\bm d_{\bm q_0}$ on the payload are small), and the two drones are required to fly towards the desired positions \eqref{fixedpositions} as in point stabilization.
The disturbances on the payload are set as
\begin{equation}
    \begin{aligned}
        \bm d_{\bm p_0} &= 
        \left\{
        \begin{aligned}
            &[0, 0, 0]^\top, N, \quad t < 7s\\
            &[5, 5, 0]^\top, N, \quad 7\le t \le 10s\\
            &[0, 0, 0]^\top, N, \quad t > 10s
        \end{aligned}
        \right.\\
        \bm d_{\bm q_0} &= 
        [0, 0, 0]^\top, N
    \end{aligned}
\end{equation}
Fig. \ref{simulation_3_dp} shows the disturbances $\bm d_{\bm p_i} \ (i=1,2)$ and their estimates  before, during, and after $\bm d_{\bm p_0}$ activation. 
Despite the unmodelled payload disturbance, DO-ESEKF maintains stable estimates of the disturbances (red lines) without significant divergence from the ground truth (black lines). This indicates the robustness of the proposed method to payload disturbances while preserving accuracy for primary estimation targets (disturbances on the UAVs and system states).
In addition,
the result in Fig. \ref{simulation_3_dp} is consistent with the results of observability analysis in Section \ref{observability} (full observability is lost when three types lumped disturbances exist).

\section{Experimental Validation}
This section supplies indoor flight experimental results to verify the
 proposed approach. 
The experimental facility consists of optitrack motion capture system (MOCAP), ground station, and quadrotor platforms, which have been used to support various research projects \cite{xu2024oscillation, xu2024force}.
System and estimation parameters in the experiment are listed in Tab. \ref{expparam}.

\begin{figure}[!h]
    \centering
    \includegraphics[width=3.5in]{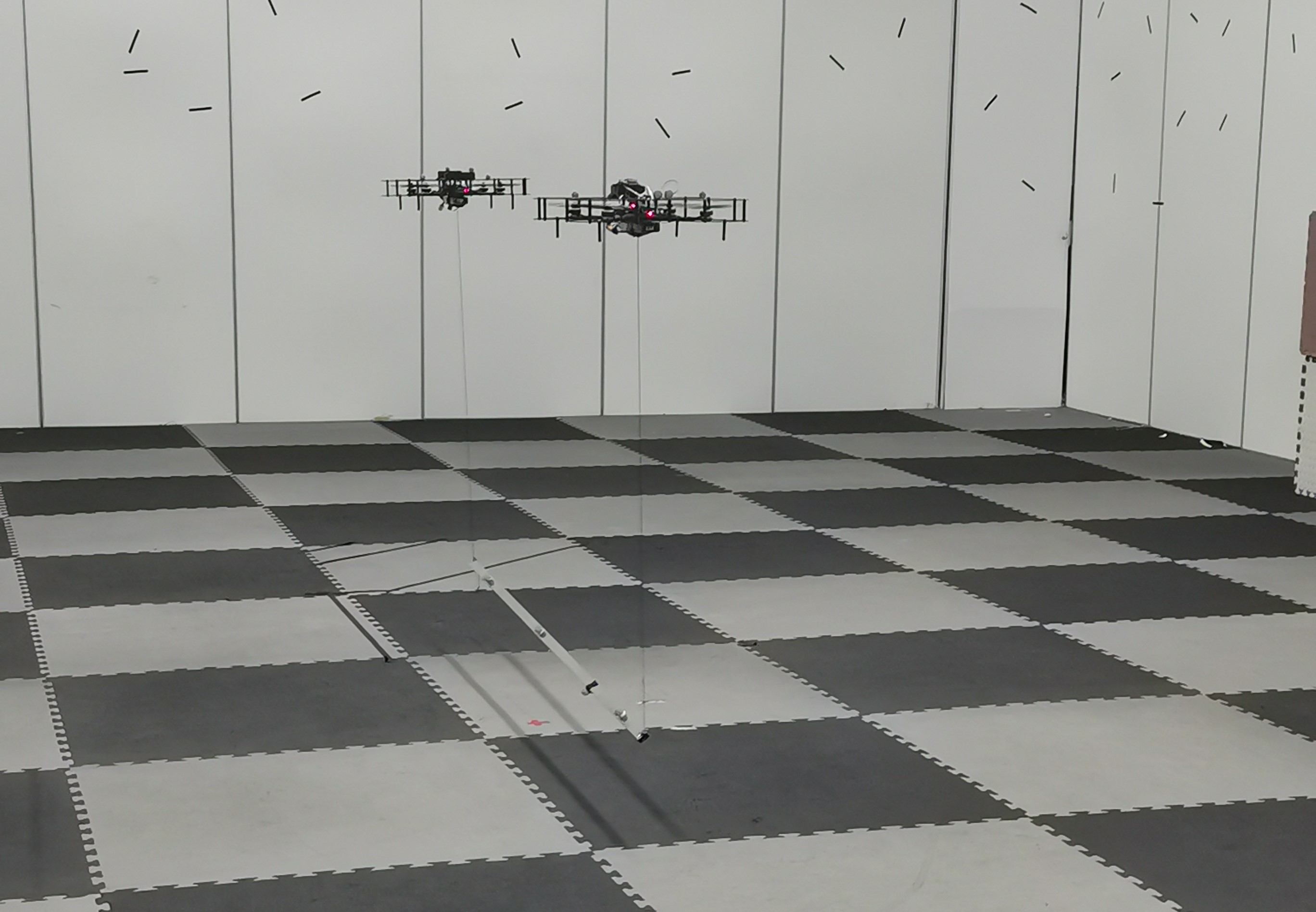}
    \caption{The experimental schematic}
    \label{exp-schematic}
\end{figure}

In the experiment, 
the two quadrotors are required to follow the desired trajectories as
\begin{equation*}
    \begin{aligned}
        \bm p_{1}^{des} (t) &= [1+0.7\sin t, 0.7\sin t\cos t, -1.4+0.1\sin t]^\top, m\\
        \bm p_{2}^{des} (t) &= [-1+0.7\sin t, 0.7\sin t\cos t, -1.4+0.15\sin t]^\top, m\\
    \end{aligned}
\end{equation*}
The experimental schematic is shown in Fig. \ref{exp-schematic},
where the poses of the quadrotors and the payload can be captured by the MOCAP via the installed markers. 
The measurements \eqref{actual_measurements} in the experiments are derived by fusing the positions of the quadrotors measured by the MOCAP and the accelerations of the quadrotors by the onboard IMUs through the common EKF \cite{px4ekf}.

\begin{figure}[!h]
    \centering
    \includegraphics[width=3.5in]{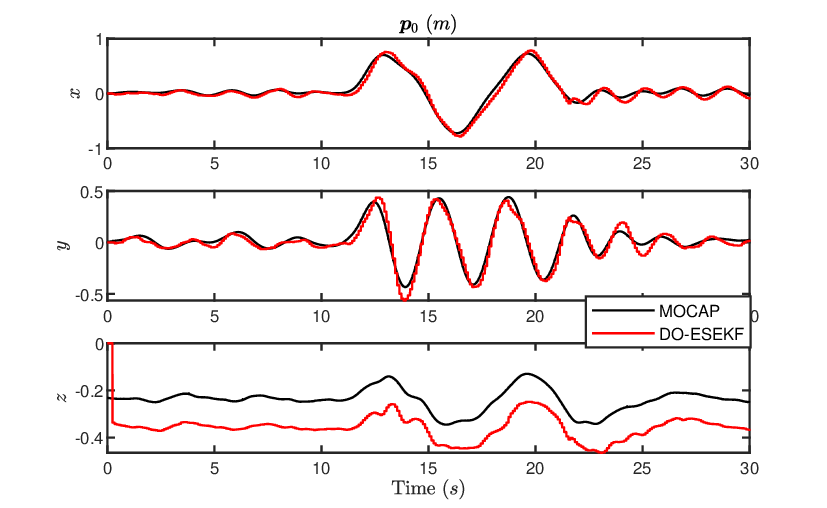}
    \caption{The estimation results of payload position in the experiment}
    \label{exp_p0}
\end{figure}

\begin{figure}[!h]
    \centering
    \includegraphics[width=3.5in]{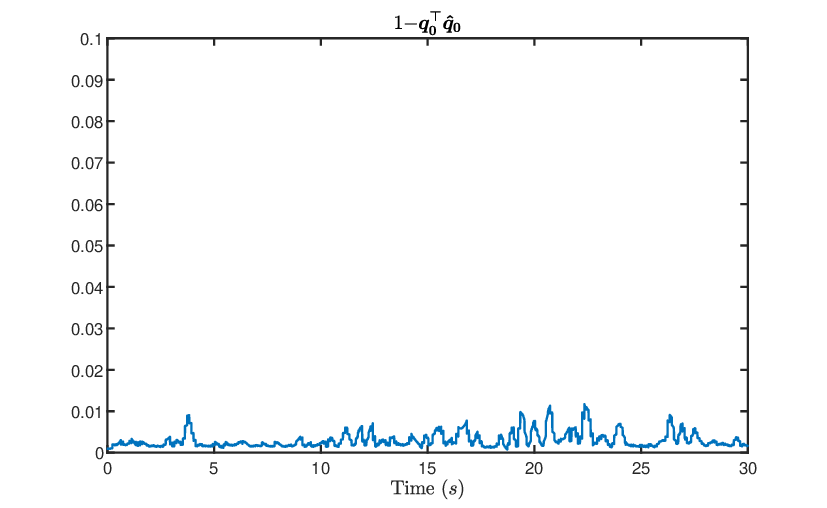}
    \caption{The estimation error of payload attitude in the experiment}
    \label{exp_q0}
\end{figure}

\begin{figure}[!h]
    \centering
    \includegraphics[width=3.5in]{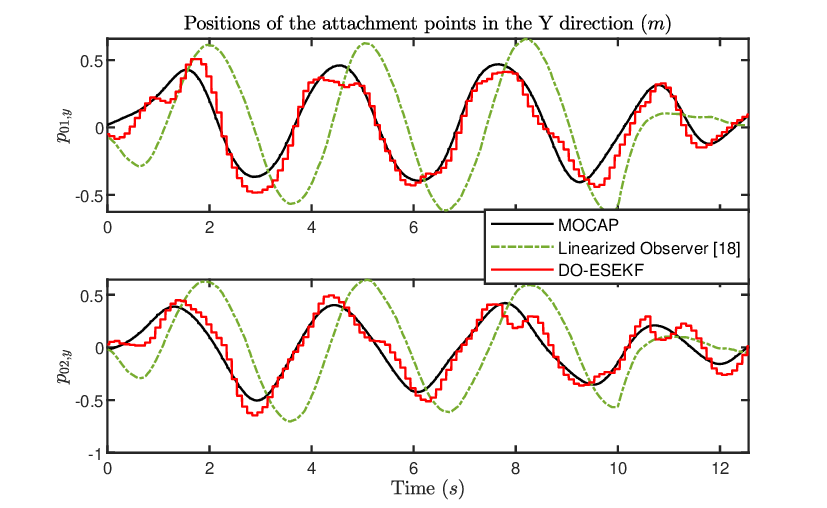}
    \caption{The comparison of payload estimation with the linearized method \cite{xu2024oscillation} in the experiment}
    \label{exp_comp}
\end{figure}

The proposed DO-based ESEKF is implemented in the ground station due to its high computational complexity.
Limited by the downlink rate of the input measurements from the drones to the ground station,
the loop rate of the proposed filter is set as 10 Hz.
 Similar to the simulation, we record the CoM position of the payload in Fig. \ref{exp_p0}.
 By comparing the DO-ESEKF estimates (red lines) of the payload CoM position with the MOCAP observations (black lines),
 it can be concluded that the proposed DO-ESEKF method  renders high-precision position estimates throughout the agile flight maneuvers, especially in the $X$ and $Y$ direction.
 Since the cable attachment points are lower than the CoM positions of the drones in the $Z$ direction in practice,
 the position estimate in the $Z$ direction in Fig. \ref{exp_p0} exhibits a fixed estimation error of approximately 10 cm,
 which can be further calibrated offline.
 To evaluate the estimation performance of the payload attitude,
 a nonlinear attitude error is formulated as 
 $$e_{\bm q_0} = 1 - \bm q_0^\top \hat{\bm q}_0$$
 where $\bm q_0$ denotes the direction of the bar axis captured by the MOCAP, 
 and $\hat{\bm q}_0$ denotes the corresponding estimate by the DO-based ESEKF.
 The nonlinear error $e_{\bm q_0}$ is depicted in Fig. \ref{exp_q0}, which remains between 0 and 0.02, indicating that the equivalent angle error is less than 2 degrees.
 This result further demonstrates the accuracy of the proposed method in pose estimation of the payload.
 Similar to the simulation in Section \ref{simulation},
 we add the method in \cite{xu2024oscillation} for comparison as shown in Fig. \ref{exp_comp},
where DO-ESEKF achieves significantly lower estimation error than the linearized observer does, demonstrating the superiority of the proposed method.
\begin{table}[!h]
	\centering
	\caption{Parameters used in the experiments}
	\resizebox{!}{!}{
		\begin{tabular}{c  c | c  c} 
			\hline 
			\hline
			\multicolumn{4}{c}{System Parameters}\\
			\hline
                \multicolumn{2}{c}{Payload} & \multicolumn{2}{c}{Drones}\\
                \hline
                $m_0$ & $0.445, kg$ & $m_1$ & $0.896, kg$\\
			\hline
                $J_0$ & $0.148, kg\cdot m^2$ & $m_2$ & $0.825, kg$\\
                \hline
                $\rho_1$ & $1, \mathrm{m}$ & $l_1$ & $1.03, m$\\
                \hline 
                $\rho_2$ & $1, \mathrm{m}$ & $l_2$ & $1.05, m$\\
                \hline
			\multicolumn{4}{c}{Estimation Parameters}\\
			\hline
			\multicolumn{2}{c}{$\bm Q_k$} & \multicolumn{2}{c}{$1 \times {10}^{0}\bm I_{12}$}\\
                \multicolumn{2}{c}{$\bm R_k$} & \multicolumn{2}{c}{$1 \times 10^{-2} \bm I_{12}$}\\
			\hline
			\hline
	\end{tabular}}\label{expparam}
\end{table}

\section{Conclusion}
 This article has established a comprehensive framework for estimating the pose of a tethered bar payload under multi-source disturbances, encompassing theoretical analysis, filter design, and experimental validation. The core advantage of our approach lies in its ability to accurately estimate the payload pose using only the UAVs' odometry data, thereby eliminating the reliance on extra sensors mounted on the payload itself. Our observability analysis provides, for the first time, a theoretical guarantee for this sensor-free estimation paradigm, confirming the system's observability under mild conditions. The proposed DO-based ESEKF effectively translates this theory into practice, delivering robust estimation performance that significantly outperforms linearized methods, especially during aggressive maneuvers far from equilibrium. Extensive simulations and real-world experiments have consistently validated the filter's effectiveness and practicality. For future work, we will investigate adaptive techniques to handle uncertainties in model parameters, further enhancing the method's robustness.

\section*{Appendix}

\subsection{Important manifolds in practice and their derivatives}\label{manifold_derivatives}

\subsubsection{Euclidean Space $\mathcal M = \mathbb R^n$}
\begin{equation}\nonumber
    \begin{aligned}
         &\bm x \boxplus \bm u = \bm x + \bm u\\
         &\bm x \boxminus \bm y = \bm x - \bm y\\
        &\bm x \oplus \bm v = \bm x + \bm v\\
       &\frac{\partial (((\bm x\boxplus \bm u)\oplus \bm v) \boxminus \bm y)}{\partial \bm u} = \bm I_n\\
        &\frac{\partial (((\bm x\boxplus \bm u)\oplus \bm v)\boxminus \bm y)}{\partial \bm v} = \bm I_n\\
    \end{aligned}
\end{equation}

\subsubsection{2-sphere $\mathcal M = \mathsf S^2$}
\begin{equation}\nonumber
    \begin{aligned}
         &\bm x \boxplus \bm u = \bm R(\bm B(\bm x) \bm u) \bm x\\
         &\bm x \boxminus \bm y = \bm B(\bm y)^\top\left( \theta \frac{\bm y^\times \bm x}{\|\bm y^\times \bm x\|}\right)\\
        &\bm x \oplus \bm v = \bm R(\bm v) \bm x\\
       &\frac{\partial (((\bm x\boxplus \bm u)\oplus \bm v) \boxminus \bm y)}{\partial \bm u} = \bm N((\bm x \boxplus \bm u) \oplus \bm v, \bm y)\bm R(\bm v)\bm O(\bm x, \bm u)\\
        &\frac{\partial (((\bm x\boxplus \bm u)\oplus \bm v)\boxminus \bm y)}{\partial \bm v} = -\bm N((\bm x\boxplus \bm u) \oplus \bm v, \bm y) \bm R(\bm v)(\bm x\boxplus \bm u)^\times \bm A(\bm v)^\top\\
    \end{aligned}
\end{equation}
where $\theta = \mathrm{atan2}(\|\bm y^\times \bm x\|, \bm y^\top \bm x)$ denotes the angle between the vector $\bm x$ and $\bm y$, $\bm R(\bm v)  \in \mathsf{SO}(3)$ denotes the rotation about the axis-angle represented by $\bm v \in \mathbb R^3$, and $\bm B(\bm x) = [\bm b_1, \bm b_2] \in \mathbb R^{3\times 2}$ denotes the two orthonormal bases lying on the tangent plane of $\bm x \in \mathsf S^2$.
The relevant matrices are defined as
\begin{equation}\nonumber
    \begin{aligned}
    \bm R (\bm v) &= \bm I_3 + \frac{\sin(\|\bm v\|)}{\|\bm v\|} \bm v^\times + \frac{1-\cos(\|\bm v\|)}{\|\bm v\|^2} (\bm v^\times)^2\\
        \bm A(\bm v) &= \bm I_3 + \frac{1-\cos(\|\bm v\|)}{\|\bm v\|^2}\bm v^\times +\frac{\|\bm v\|-\sin(\|\bm v\|)}{\|\bm v\|^3}(\bm v^\times)^2\\
        \bm N(\bm x, \bm y)&= \bm B(\bm y)^\top \left(\frac{\theta}{\|\bm y^\times \bm x\|} \bm y^\times + (\bm y^\times \bm x)\bm P(\bm x, \bm y)\right)\\
        \bm O(\bm x, \bm u) &= -\bm R(\bm B(\bm x) \bm u) \bm x^\times \bm A(\bm B(\bm x) \bm u)^\top \bm B(\bm x)\\
        \bm P(\bm x, \bm y) &= \frac{-\bm y^\top\bm x \|\bm y^\times \bm x\| + \theta}{\|\bm y^\times \bm x\|^3} \bm x^\top (\bm y^\times)^2-\bm y^\top
    \end{aligned}
\end{equation}
Following \cite{he2023symbolic}, the expression of the matrix $\bm B(\bm x)$  is adopted as
\begin{equation*}
    \begin{aligned}
        \bm B (\bm x) &= \bm R_1(\bm x) [\bm e_2, \ \bm e_3]\\
        \bm R_1(\bm x) &= \bm R \left(\frac{\bm e_1^\times \bm x}{\|\bm e_1^\times \bm x\|} \mathrm{atan2} \left(\|\bm e_1^\times \bm x\|, \bm e_1^\top \bm x\right) \right)
    \end{aligned}
\end{equation*}
where $\bm e_i \ (i=1,2,3)$ are 3-dimension unit vectors.

\subsubsection{Tangent bundle of 2-sphere $\mathcal M = \mathsf{TS}^2$}
\begin{equation}\nonumber
    \begin{aligned}
         &\bm x \boxplus \bm u =
         \begin{bmatrix}
             \bm x_1 \boxplus \bm u_1\\
             \bm x_2 \boxplus \bm u_2
         \end{bmatrix}=
         \begin{bmatrix}
             \bm R(\bm B(\bm x_1) \bm u_1) \bm x_1\\
             \bm x_2 + \bm R(\bm B(\bm x_1) \bm u_1) \bm B (\bm x_1) \bm u_2
         \end{bmatrix}\\
         &\bm x \boxminus \bm y =
         \begin{bmatrix}
             \bm x_1 \boxminus \bm y_1\\
             \bm x_2 \boxminus \bm y_2
         \end{bmatrix}
         = \begin{bmatrix}
             \bm B(\bm y_1)^\top\left( \theta \frac{\bm y_1^\times \bm x_1}{\|\bm y_1^\times \bm x_1\|}\right)\\
             \bm B(\bm y_1)^\top\left( \bm R \left(-\theta \frac{\bm y_1^\times \bm x_1}{\|\bm y_1^\times \bm x_1\|}\right)\bm x_2 - \bm y_2\right)\\
         \end{bmatrix}\\
        &\bm x \oplus \bm v = 
        \begin{bmatrix}
            \bm x_1 \oplus \bm v_1\\
            \bm x_2 \oplus \bm v_2
        \end{bmatrix} = 
        \begin{bmatrix}
            \bm R(\bm v_1) \bm x_1\\
            \bm x_2 + \bm R (\bm v_1) \bm v_2
        \end{bmatrix}\\
       &\frac{\partial (((\bm x\boxplus \bm u)\oplus \bm v) \boxminus \bm y)}{\partial \bm u} = \bm Q((\bm x \boxplus \bm u) \oplus \bm v, \bm y) \bm S(\bm v)\bm T(\bm x, \bm u)\\
        &\frac{\partial (((\bm x\boxplus \bm u)\oplus \bm v)\boxminus \bm y)}{\partial \bm v} = \bm Q((\bm x\boxplus \bm u) \oplus \bm v, \bm y) \bm U(\bm x\boxplus \bm u, \bm v)\\
    \end{aligned}
\end{equation}
where $(\cdot)_1$ denotes the component on the manifold $\mathsf S^2$, and $(\cdot)_2$ denotes the component on its corresponding tangent space.
$\theta = \mathrm{atan2}(\|\bm y_1^\times \bm x_1\|, \bm y_1^\top \bm x_1)$ denotes the angle between the vector $\bm x_1$ and $\bm y_1$.
The relevant matrices are defined as
\begin{equation*}
    \begin{aligned}
    \bm Q (\bm x, \bm y) &= \frac{\partial (\bm x \boxminus \bm y)}{\partial \bm x}
    = \begin{bmatrix}
        \frac{\partial (\bm x_1 \boxminus \bm y_1)}{\partial \bm x_1} & \frac{\partial (\bm x_1 \boxminus \bm y_1)}{\partial \bm x_2}\\
        \frac{\partial (\bm x_2 \boxminus \bm y_2)}{\partial \bm x_1} & \frac{\partial (\bm x_2 \boxminus \bm y_2)}{\partial \bm x_2}
    \end{bmatrix}\\
    &=\begin{bmatrix}
        \bm B(\bm y_1)^\top \left(\frac{\theta}{\|\bm y_1^\times \bm x_1\|} \bm y_1^\times + (\bm y_1^\times \bm x_1)\bm P(\bm x_1, \bm y_1)\right) & \bm 0_{2\times 3}\\
        \bm B(\bm y_1)^\top \bm R\left( -\theta \frac{\bm y_1^\times \bm x_1}{\|\bm y_1^\times \bm x_1\|}\right) \bm x_2^\times \bm A \left(-\theta \frac{\bm y_1^\times \bm x_1}{\|\bm y_1^\times \bm x_1\|}\right)^\top \left(\frac{\theta}{\|\bm y_1^\times \bm x_1\|} \bm y_1^\times + (\bm y_1^\times \bm x_1)  \bm P(\bm x_1, \bm y_1)\right) &
        \bm B (\bm y_1)^\top \bm R \left(-\theta \frac{\bm y_1^\times \bm x_1}{\|\bm y_1^\times \bm x_1\|}\right)
    \end{bmatrix}\\
    \bm S(\bm v) &= \frac{\partial (\bm x \oplus \bm v)}{\partial \bm x}
        = \begin{bmatrix}
            \frac{\partial (\bm x_1 \oplus \bm v_1)}{\partial \bm x_1} &
            \frac{\partial (\bm x_1 \oplus \bm v_1)}{\partial \bm x_2}\\
            \frac{\partial (\bm x_2 \oplus \bm v_2)}{\partial \bm x_1} &
            \frac{\partial (\bm x_2 \oplus \bm v_2)}{\partial \bm x_2}
        \end{bmatrix}\\
    &= \begin{bmatrix}
            \bm R(\bm v_1) & \bm 0_{3\times 3}\\
            \bm 0_{3\times 3} & \bm I_3
        \end{bmatrix}\\
        \bm T (\bm x, \bm u) &= \frac{\partial (\bm x \boxplus \bm u)}{\partial \bm u}
        = \begin{bmatrix}
            \frac{\partial (\bm x_1 \boxplus \bm u_1)}{\partial \bm u_1} & \frac{\partial (\bm x_1 \boxplus \bm u_1)}{\partial \bm u_2}\\
            \frac{\partial (\bm x_2 \boxplus \bm u_2)}{\partial \bm u_1} &
            \frac{\partial (\bm x_2 \boxplus \bm u_2)}{\partial \bm u_2}
        \end{bmatrix}\\
        &= \begin{bmatrix}
            -\bm R(\bm B(\bm x_1) \bm u_1) \bm x_1^\times \bm A(\bm B(\bm x_1) \bm u_1)^\top \bm B (\bm x_1) & \bm 0_{3\times 2}\\
            -\bm R(\bm B(\bm x_1) \bm u_1) (\bm B(\bm x_1)\bm u_2)^\times \bm A(\bm B(\bm x_1)\bm u_1)^\times \bm B (\bm x_1) & \bm R(\bm B (\bm x_1) \bm u_1) \bm B(\bm x_1) 
        \end{bmatrix}\\
        \bm U (\bm x, \bm v) &=
        \frac{\partial (\bm x \oplus \bm v)}{\partial \bm v}
        = \begin{bmatrix}
            \frac{\partial (\bm x_1 \oplus \bm v_1)}{\partial \bm v_1} &
            \frac{\partial (\bm x_1 \oplus \bm v_1)}{\partial \bm v_2}\\
            \frac{\partial (\bm x_2 \oplus \bm v_2)}{\partial \bm v_1} &
            \frac{\partial (\bm x_2 \oplus \bm v_2)}{\partial \bm v_2}
        \end{bmatrix}\\
        &= \begin{bmatrix}
            -\bm R (\bm v_1) \bm x_1^\times \bm A (\bm v_1)^\top &
            \bm 0_{3\times 3}\\
            -\bm R (\bm v_1) \bm v_2^\times \bm A (\bm v_1)^\top &
            \bm R (\bm v_1)
        \end{bmatrix}
    \end{aligned}
\end{equation*}
Since only $\bm Q(\bm x, \bm x)$ is used in the partial differentiations, the corresponding matrix is simplified as
\begin{equation*}
    \begin{aligned}
        \bm Q(\bm x, \bm x) &= 
        \begin{bmatrix}
            \bm B(\bm x_1)^\top \bm x_1^\times & \bm 0_{2\times 3}\\
            \bm B(\bm x_1)^\top \bm x_2^\times \bm x_1^\times & \bm B(\bm x_1)^\top 
        \end{bmatrix}
    \end{aligned}
\end{equation*}

\subsection{Manifold-specific partial differentiations}
The manifold-specific partial differentiations $\bm G_{(\cdot)}$ and $\bm J_{(\cdot)}$ can be computed as
\begin{equation*}
    \begin{aligned}
        \bm G_{\bm x_\tau} &= \left.\frac{\partial (((\bm x \boxplus \bm u)\oplus \bm v)\boxminus \bm y)}{\partial \bm u}\right|_{\bm x = \bm x_{\tau}^+, \bm u = \bm 0, \bm v = \Delta t \bm f_d(\bm x_\tau^+, \bm u_\tau, \bm 0), \bm y = \bm x_{\tau+1}^-}\\
        \bm G_{\bm f_\tau} &= \left.\frac{\partial((\bm x \oplus \bm v)\boxminus \bm y)}{\partial \bm v}\right|_{\bm x = \bm x_\tau^+, \bm v = \Delta t \bm f_d(\bm x_\tau^+, \bm u_\tau, \bm 0), \bm y = \bm x_{\tau+1}^-}\\
        \bm J_{\tau} &= \left.\frac{\partial ((\bm x \boxplus \bm u)\boxminus \bm y)}{\partial \bm u}\right|_{\bm x = \bm x_\tau^-, \bm u = \delta \bm x_\tau^o, \bm y = \bm x_\tau^+}\\
    \end{aligned}
\end{equation*}
Based on the manifold derivatives in Appendix A, 
the relevant partial differentiations for states evolving on $\mathbb R^n$ can be computed as
\begin{equation*}
    \begin{aligned}
        \bm G_{\bm x_\tau}^{\bm p_0, \bm v_0} &= \bm G_{\bm x_\tau}^{\bm d_{\bm p_1}, \bm d_{\bm p_2}} = \bm I_6,\\
        \bm G_{\bm f_\tau}^{\bm p_0, \bm v_0} &= 
        \bm G_{\bm x_\tau}^{\bm d_{\bm p_1}, \bm d_{\bm p_2}} 
        = \bm I_6,\\
        \bm J_\tau^{\bm p_0, \bm v_0} &= 
        \bm J_\tau^{\bm d_{\bm p_1}, \bm d_{\bm p_2}}
        = \bm I_6,
    \end{aligned}
\end{equation*}
and the relevant partial differentiations for states evolving on $\mathsf{TS}^2$ can be computed as
\begin{equation*}
    \begin{aligned}
         \bm G_{\bm x_\tau}^{\bm q_i, \bm \omega_i} &=
        \left.\bm Q(\bm y, \bm y) \bm S(\bm v) \bm T(\bm x, \bm 0)\right|_{\bm x = [ \bm q_{i,\tau}^+; \bm \omega_{i,\tau}^+], \ \bm v = \Delta t [\bm \omega_{i,\tau}^+; \dot{\bm \omega}_{i,\tau}^+], \ \bm y = [\bm q_{i,\tau+1}^-; \bm \omega_{i,\tau+1}^-]}\\
        &= \begin{bmatrix}
            \bm B (\bm q_{i,\tau+1}^-)^\top (\bm q_{i,\tau+1}^-)^\times & \bm 0_{2\times 3}\\
            \bm B (\bm q_{i,\tau+1}^-)^\top (\bm \omega_{i,\tau+1}^-)^\times (\bm q_{i,\tau+1}^-)^\times & \bm B (\bm q_{i,\tau+1}^-)^\top
            \end{bmatrix}
            \begin{bmatrix}
                \bm R(\Delta t \bm \omega_{i,\tau}^+) & \bm 0_{3\times 3}\\
                \bm 0_{3\times 3} & \bm I_3
            \end{bmatrix}
            \begin{bmatrix}
                -(\bm q_{i,\tau}^+)^\times \bm B(\bm q_{i,\tau}^+) & \bm 0_{3\times 2}\\
                \bm 0_{3\times 2} & \bm B(\bm q_{i,\tau}^+)
            \end{bmatrix}\\
            &= \begin{bmatrix}
                -\bm B(\bm q_{i,\tau+1}^-)^\top \bm R(\Delta t \bm \omega_{i,\tau}^+) ((\bm q_{i,\tau}^+)^\times)^2 \bm B (\bm q_{i,\tau}^+) & \bm 0_{2\times 2}\\
                \bm 0_{2\times 2} & \bm B(\bm q_{i,\tau+1}^-)^\top \bm B (\bm q_{i,\tau}^+)
            \end{bmatrix}, \quad i=0,1,2\\
    \end{aligned}
\end{equation*}

\begin{equation*}
    \begin{aligned}
            \bm G_{\bm f_\tau}^{\bm q_i, \bm \omega_i} &=
            \left. \bm Q (\bm y, \bm y) \bm U (\bm x, \bm v)\right|_{\bm x = [\bm q_{i,\tau}^+; \bm \omega_{i,\tau}^+], \ \bm v = \Delta t [\bm \omega_{i,\tau}^+; \dot{\bm \omega}_{i,\tau}^+], \  \bm y = [\bm q_{i,\tau+1}^-; \bm \omega_{i,\tau+1}^-]}\\
            &= \begin{bmatrix}
            \bm B (\bm q_{i,\tau+1}^-)^\top (\bm q_{i,\tau+1}^-)^\times & \bm 0_{2\times 3}\\
            \bm B (\bm q_{i,\tau+1}^-)^\top (\bm \omega_{i,\tau+1}^-)^\times (\bm q_{i,\tau+1}^-)^\times & \bm B (\bm q_{i,\tau+1}^-)^\top
            \end{bmatrix}
            \begin{bmatrix}
                -\bm R(\Delta t \bm \omega_{i,\tau}^+) (\bm q_{i,\tau}^+)^\times \bm A (\Delta t \bm \omega_{i,\tau}^+)^\top & \bm 0_{3\times 3}\\
                -\bm R(\Delta t \bm \omega_{i,\tau}^+) (\Delta t \dot{\bm \omega}_{i,\tau}^+)^\times \bm A (\Delta t \bm \omega_{i,\tau}^+)^\top & \bm R(\Delta t \bm \omega_{i,\tau}^+)
            \end{bmatrix}\\
            &= 
            \begin{bmatrix}
                -\bm B(\bm q_{i,\tau+1}^-)^\top \bm R(\Delta t \bm \omega_{i,\tau}^+) ((\bm q_{i,\tau}^+)^\times)^2 \bm A (\Delta t \bm \omega_{i,\tau}^+)^\top & \bm 0_{2\times 3}\\
                -\bm B(\bm q_{i,\tau+1}^-)^\top \bm R(\Delta t \bm \omega_{i,\tau}^+) (\Delta t \dot{\bm \omega}_{i,\tau}^+)^\times  \bm A (\Delta t \bm \omega_{i,\tau}^+)^\top &
                \bm B(\bm q_{i,\tau+1}^-)^\top \bm R (\Delta t \bm \omega_{i,\tau}^+)
            \end{bmatrix}, \quad i=0,1,2
    \end{aligned}
\end{equation*}

\begin{equation*}
\begin{aligned}
    \bm J_\tau^{\bm q_i, \bm \omega_i} &= 
    \left.\bm Q(\bm y, \bm y) \bm S(\bm 0) \bm T(\bm x, \bm u)\right|_{\bm x = [\bm q_{i,\tau}^-; \bm \omega_{i,\tau}^-], \ \bm u = [\delta \bm q_{i,\tau}^o; \delta \bm \omega_{i,\tau}^o], \ \bm y = [\bm q_{i,\tau}^+; \bm \omega_{i,\tau}^+]}\\
    &= \begin{bmatrix}
        \bm B(\bm q_{i,\tau}^+)^\top (\bm q_{i,\tau}^+)^\times & \bm 0_{2\times 3}\\
        \bm B(\bm q_{i,\tau}^+)^\top (\bm \omega_{i,\tau}^+)^\times (\bm q_{i,\tau}^+)^\times & \bm B(\bm q_{i,\tau}^+)^\top
    \end{bmatrix}\\
    &\quad \cdot
    \begin{bmatrix}
        -\bm R( \bm B(\bm q_{i,\tau}^-) \delta \bm q_{i,\tau}^o) (\bm q_{i,\tau}^-)^\times \bm A ( \bm B(\bm q_{i,\tau}^-) \delta \bm q_{i,\tau}^o)^\top \bm B(\bm q_{i,\tau}^-) & \bm 0_{3\times 2}\\
        -\bm R( \bm B(\bm q_{i,\tau}^-) \delta \bm q_{i,\tau}^o)  (\bm B(\bm q_{i,\tau}^-) \delta \bm \omega_{i,\tau}^o)^\times \bm A( \bm B(\bm q_{i,\tau}^-) \delta \bm q_{i,\tau}^o)^\top \bm B(\bm q_{i,\tau}^-) &
        \bm R( \bm B(\bm q_{i,\tau}^-) \delta \bm q_{i,\tau}^o) \bm B(\bm q_{i,\tau}^-)
    \end{bmatrix}\\
    &= 
    \begin{bmatrix}
        -\bm B(\bm q_{i,\tau}^+)^\top \bm R ( \bm B(\bm q_{i,\tau}^-) \delta \bm q_{i,\tau}^o) ((\bm q_{i,\tau}^-)^\times)^2 \bm A ( \bm B(\bm q_{i,\tau}^-) \delta \bm q_{i,\tau}^o)^\top \bm B(\bm q_{i,\tau}^-) & \bm 0_{2\times 2}\\
        -\bm B(\bm q_{i,\tau}^+)^\top \bm R ( \bm B(\bm q_{i,\tau}^-) \delta \bm q_{i,\tau}^o)  (\bm B(\bm q_{i,\tau}^-) \delta \bm \omega_{i,\tau}^o)^\times  \bm A ( \bm B(\bm q_{i,\tau}^-) \delta \bm q_{i,\tau}^o)^\top \bm B (\bm q_{i,\tau}^-) &
        \bm B(\bm q_{i,\tau}^+)^\top \bm R ( \bm B(\bm q_{i,\tau}^-) \delta \bm q_{i,\tau}^o) \bm B(\bm q_{i,\tau}^-)
    \end{bmatrix}\\
    i&= 0,1,2
\end{aligned}
\end{equation*}
where $(\cdot)_\tau^+$ denotes the posterior estimate at the step $\tau$ and $(\cdot)_{\tau+1}^-$ denotes the prior estimate at the step $\tau+1$.
The variables $\delta \bm q_{i,\tau}^o = \bm q_{i,\tau}^+ \boxminus \bm q_{i,\tau}^-$ and
    $ \delta \bm \omega_{i,\tau}^o = \bm \omega_{i,\tau}^+ \boxminus \bm \omega_{i,\tau}^-$ are used for substitutions.
In addition, the equations $\bm q_{i,\tau+1}^- = \bm R (\Delta t \bm \omega_{i,\tau}^+) \bm q_{i,\tau}^+$ and $\bm \omega_{i,\tau+1}^-=\bm \omega_{i,\tau}^+ + \bm R(\Delta t \bm \omega_{i,\tau}^+) \Delta t \dot{\bm \omega}_{i,\tau}^+$ have been used to simplify the above partial differentiations.

\subsection{Detailed formulation of compact symbols}
For the compact form in \eqref{compact_form}, the expression of the matrix $\bm M$ is 
\begin{equation*}
    \begin{aligned}
\bm{M}= \begin{bmatrix}
	m_T\bm{I}_3&		\sum\limits_{i=1}^2{\left( -1 \right)^i m_i \rho _i\bm{q}_{0}^{\times}}&		m_1l_1\bm{q}_{1}^{\times}&		m_2l_2\bm{q}_{2}^{\times}\\
	\sum\limits_{i=1}^2{\left( -1 \right) ^{i+1}m_i\rho _i\bm{q}_{0}^{\times}}&		\bar{J}_0\bm{I}_3&		m_1\rho _1l_1\bm{q}_{0}^{\times}\bm{q}_{1}^{\times}&		-m_2\rho _2l_2\bm{q}_{0}^{\times}\bm{q}_{2}^{\times}\\
	-m_1\bm{q}_{1}^{\times}&		m_1\rho _1\bm{q}_{1}^{\times}\bm{q}_{0}^{\times}&		m_1l_1\bm{I}_3&		\bm{0}_{3\times 3}\\
	-m_2\bm{q}_{2}^{\times}&		-m_2\rho _2\bm{q}_{2}^{\times}\bm{q}_{0}^{\times}&		\bm{0}_{3\times 3}&		m_2l_2\bm{I}_3\\
\end{bmatrix}  
    \end{aligned}
\end{equation*}
and the expression of the vector $\bm F$ is
\begin{equation*}
    \begin{aligned}
\bm{F}=\left[ \begin{array}{c}
	-\sum\limits_{i=1}^2{m_i\left[ l_i\left\| \bm{\omega }_i \right\| ^2\bm{q}_i+\left( -1 \right) ^i\rho _i\left\| \bm{\omega }_0 \right\| ^2\bm{q}_0 \right]}+m_Tg\bm{e}_3+\sum\limits_{i=1}^2{(\bm{u}_i+\bm{d}_{\bm{p}_i})}+ \bm w_{\bm p_0}\\
	\sum_{i=1}^2{\left( -1 \right) ^{i+1}\rho _i\bm{q}_{0}^{\times}\left( m_ig\bm{e}_3-m_il_i\left\| \bm{\omega }_i \right\| ^2\bm{q}_i+\bm{u}_i+\bm{d}_{\bm{p}_i} \right)}+\bm{q}_{0}^{\times}\bm{w}_{\bm{q}_0}\\
	-\bm{q}_{1}^{\times}\left( m_1g\bm{e}_3+m_1\rho _1\left\| \bm{\omega }_0 \right\| ^2\bm{q}_0+\bm{u}_1+\bm{d}_{\bm{p}_1} \right)\\
	-\bm{q}_{2}^{\times}\left( m_2g\bm{e}_3-m_2\rho _2\left\| \bm{\omega }_0 \right\| ^2\bm{q}_0+\bm{u}_2+\bm{d}_{\bm{p}_2} \right)\\
\end{array} \right] 
    \end{aligned}
\end{equation*}
\subsection{System-specific partial differentiations}

For $\bm x$ in \eqref{x}, $\bm u$ in \eqref{control_input}, and $\bm f_d$ in \eqref{canonical},
$\left.\frac{\partial \bm f_d(\bm x \boxplus \delta \bm x, \bm u, \bm 0)}{\partial \delta \bm x}\right|_{\delta \bm x = \bm 0}$ can be derived as
\begin{equation}\label{f_d}
    \begin{aligned}
        \left.\frac{\partial \bm f_d(\bm x \boxplus \delta \bm x, \bm u, \bm 0)}{\partial \delta \bm x}\right|_{\delta \bm x = \bm 0}
        = 
\left. \frac{\partial \bm{f}_d\left( \bm{z},\bm{u},\bm{0} \right)}{\partial \bm{z}} \right|_{\bm{z}=\bm{x}}\cdot \left. \frac{\partial \left( \bm{x}\boxplus \delta \bm{x} \right)}{\partial \delta \bm{x}} \right|_{\delta \bm{x}=\bm 0}
    \end{aligned}
\end{equation}
The components on the right hand-side of the equation \eqref{f_d} are calculated as
\begin{equation}\label{x_boxplus}
    \begin{aligned}
    \left.\frac{\partial \bm f(\bm z, \bm u, \bm 0)}{\partial \bm z}\right|_{\bm z = \bm x} &= \left.
    \begin{bmatrix}
        \bm 0_{3\times 3} & \bm I_3 & \bm 0_{3\times 24}\\
        \hdashline
        & (\partial \bm M^{-1} \bm F/ \partial \bm z)_{1:3} &\\
        \hdashline
        \bm 0_{3\times 9} & \bm I_3 & \bm 0_{3\times 18}\\
        \hdashline
        & (\partial \bm M^{-1} \bm F/ \partial \bm z)_{4:6} &\\
        \hdashline
        \bm 0_{3\times 15} & \bm I_3 & \bm 0_{3\times 12}\\
        \hdashline
        & (\partial \bm M^{-1} \bm F/ \partial \bm z)_{7:9} &\\
        \hdashline
        \bm 0_{3\times 21} & \bm I_3 & \bm 0_{3\times 6}\\
        \hdashline
        & (\partial \bm M^{-1} \bm F/ \partial \bm z)_{10:12} &\\
        \hdashline
        & \bm 0_{6 \times 30} &
        \end{bmatrix}\right|_{\bm z = \bm x}\\
        \left. \frac{\partial \left( \bm{x}\boxplus \delta \bm{x} \right)}{\partial \delta \bm{x}} \right|_{\delta \bm{x}=\bm 0} &= \mathrm{blkdiag}\left( \bm I_6, -\bm q_0^\times \bm B(\bm q_0), \bm B(\bm q_0), -\bm q_1^\times \bm B(\bm q_1), \bm B(\bm q_1), -\bm q_2^\times \bm B(\bm q_2), \bm B(\bm q_2), \bm I_6\right)
    \end{aligned}
\end{equation}
and $\partial \bm M^{-1}\bm F/\partial \bm z$ is further calculated as
\begin{equation*}
    \begin{aligned}
        \frac{\partial \bm M^{-1}\bm F}{\partial \bm z} &=
        \frac{\partial \bm M^{-1}}{\partial \bm z} \bm F + \bm M^{-1} \frac{\partial \bm F}{\partial \bm z}\\
        &=
        -\bm M^{-1} \left.\frac{\partial \bm M(\bm z)\bm \iota}{\partial \bm z}\right|_{\bm \iota= \bm M^{-1} \bm F} + \bm M^{-1} \frac{\partial \bm F}{\partial \bm z}\\
    \end{aligned}
\end{equation*}
Here 
\begin{small}
\begin{equation*}
    \begin{aligned}
        &\left.\frac{\partial \bm M(\bm z) \bm \iota}{\partial \bm z}\right|_{\bm \iota = \bm M^{-1}\bm F} = 
        \begin{bmatrix}
            \bm 0_{3\times 6} & \sum\limits_{i=1}^2{(-1)^{i+1}m_i \rho_i \bm \iota_{4:6}^\times} & \bm 0_{3\times 3} & -m_1 l_1 \bm \iota_{7:9}^\times & \bm 0_{3\times 3} & - m_2 l_2 \bm \iota_{10:12}^\times & \bm 0_{3\times 9}\\
            \bm 0_{3\times 6} & \bm \Xi & \bm 0_{3\times 3} & -m_1 \rho_1 l_1 \bm q_0^\times \bm \iota_{7:9}^\times & \bm 0_{3\times 3} & m_2 \rho_2 l_2 \bm q_0^\times \bm \iota_{10:12}^\times & \bm 0_{3\times 9}\\
            \bm 0_{3\times 6} & -m_1 \rho_1 \bm q_1^\times \bm \iota_{4:6}^\times & \bm 0_{3\times 3} & m_1 \bm \iota_{1:3}^\times - m_1 \rho_1 (\bm q_0^\times \bm \iota_{4:6})^\times & \bm 0_{3\times 3} & \bm 0_{3\times 3} & \bm 0_{3\times 9}\\
            \bm 0_{3\times 6} & m_2 \rho_2 \bm q_2^\times \bm \iota_{4:6}^\times & \bm 0_{3\times 3} & \bm 0_{3\times 3} & \bm 0_{3\times 3} & m_2 \bm \iota_{1:3}^\times+m_2 \rho_2 (\bm q_0^\times \bm \iota_{4:6})^\times & \bm 0_{3\times 9}
        \end{bmatrix}\\
        &\frac{\partial \bm F}{\partial \bm z} = \left[
        \begin{array}{ccccc}
            \bm 0_{3\times 6} & \sum\limits_{i=1}^2{(-1)^{i+1}m_i \rho_i \|\bm \omega_0\|^2 \bm I_3} &
            \sum\limits_{i=1}^2{(-1)^{i+1}2m_i \rho_i \bm q_0 \bm \omega_0^\top} &
            -m_1 l_1 \|\bm \omega_1\|^2 \bm I_3 &
            -2m_1 l_1 \bm q_1 \bm \omega_1^\top\\
          \bm 0_{3\times 6} & \sum\limits_{i=1}^2{(-1)^i \rho_i (\bm \chi_i - m_i l_i \|\bm \omega_i\|^2 \bm q_i)^\times} & \bm 0_{3\times 3} &
          -m_1 l_1 \rho_1 \|\bm \omega_1 \|^2 \bm q_0^\times & -2 m_1l_1\rho_1\bm q_0^\times \bm q_1 \bm \omega_1^\top\\
          \bm 0_{3\times 6} & -m_1 \rho_1 \|\bm \omega_0\|^2 \bm q_1^\times & -2 m_1 \rho_1 \bm q_1^\times \bm q_0 \bm \omega_0^\top & (\bm \chi_1 + m_1\rho_1 \|\bm \omega_0\|^2 \bm q_0)^\times & \bm 0_{3\times 3}\\
          \bm 0_{3\times 6} & m_2\rho_2\|\bm \omega_0\|^2 \bm q_2^\times & 2 m_2 \rho_2 \bm q_2^\times \bm q_0 \bm \omega_0^\top & \bm 0_{3\times 3} & \bm 0_{3\times 3}
        \end{array}
        \right.\\
        &\quad \quad \left.
        \begin{array}{cccc}
           -m_2 l_2 \|\bm \omega_2\|^2 \bm I_3  &
           -2 m_2 l_2 \bm q_2 \bm \omega_2^\top &
           \bm I_3 & \bm I_3\\
           m_2 l_2 \rho_2\| \bm \omega_2\|^2 \bm q_0^\times &
           2 m_2 l_2 \rho_2 \bm q_0^\times \bm q_2 \bm \omega_2^\top & \rho_1 \bm q_0^\times & -\rho_2 \bm q_0^\times\\
            \bm 0_{3\times 3} & \bm 0_{3\times 3} & -\bm q_1^\times & \bm 0_{3\times 3}\\
            (\bm \chi_2 - m_2 \rho_2 \|\bm \omega_0\|^2 \bm q_0)^\times & \bm 0_{3\times 3} & \bm 0_{3\times 3} & - \bm q_2^\times
        \end{array}
        \right]
    \end{aligned}
\end{equation*}
\end{small}
where $\bm \Xi = \sum\limits_{i=1}^2{(-1)^{i}m_i \rho_i \bm \iota_{1:3}^\times}-m_1 \rho_1 l_1 (\bm q_1^\times \bm \iota_{7:9})^\times + m_2 \rho_2 l_2 (\bm q_2^\times \bm \iota_{10:12})^\times$ and $\bm \chi_i = m_i g \bm e_3 + \bm u_i + \bm d_{\bm p_i} \ (i=1,2)$ are used for substitutions.

\vspace{0.5cm}
In addition, $\left.\frac{\partial\bm f_d(\bm x, \bm u, \bm w)}{\partial \bm w}\right|_{\bm w = \bm 0}$ can be calculated as
\begin{equation*}
    \begin{aligned}
        \left.\frac{\partial \bm f(\bm x, \bm u, \bm w)}{\partial \bm w}\right|_{\bm w = \bm 0} &= \left.
    \begin{bmatrix}
        &\bm 0_{3\times 12}&\\
        &(\partial \bm M^{-1} \bm F/ \partial \bm w)_{1:3}&\\
        &\bm 0_{3\times 12}&\\
        &(\partial \bm M^{-1} \bm F/ \partial \bm w)_{4:6}&\\
        &\bm 0_{3\times 12}&\\
        &(\partial \bm M^{-1} \bm F/ \partial \bm w)_{7:9}&\\
        &\bm 0_{3\times 12}&\\
        &(\partial \bm M^{-1} \bm F/ \partial \bm w)_{10:12}&\\
        \hdashline
        \bm 0_{3\times 6} & \bm I_3 & \bm 0_{3 \times 3}\\
        \bm 0_{3\times 6} & \bm 0_{3\times3} & \bm I_{3}\\
        \end{bmatrix}\right|_{\bm w = \bm 0}
    \end{aligned}
\end{equation*}
where $\partial \bm M^{-1}\bm F/\partial \bm w$ is further calculated as
\begin{equation*}
    \begin{aligned}
        \frac{\partial \bm M^{-1}\bm F}{\partial \bm w} &= 
        \bm M^{-1} \frac{\partial \bm F}{\partial \bm w}\\
        &= 
        \bm M^{-1} \mathrm{blkdiag} \left(\bm I_3, \bm q_0^\times, \bm 0_{6\times6}\right)
    \end{aligned}
\end{equation*}
Finally, 
$\left.\frac{\partial \bm h(\bm x \boxplus \delta \bm x)}{\partial \delta \bm x}\right|_{\delta \bm x = \bm 0}$ is derived as
\begin{equation*}
    \begin{aligned}
        \left.\frac{\partial \bm h(\bm x \boxplus \delta \bm x)}{\partial \delta \bm x}\right|_{\delta \bm x = \bm 0} &=
        \left.\frac{\partial\bm h(\bm z)}{\partial \bm z}\right|_{\bm z = \bm x} \cdot \left.\frac{\partial (\bm x\boxplus \delta \bm x)}{\partial \delta \bm x}\right|_{\delta \bm x = \bm 0}
    \end{aligned}
\end{equation*}
where $\partial (\bm x \boxplus \delta \bm x)/\partial \delta \bm x|_{\delta \bm x = \bm 0}$ has been calculated in \eqref{x_boxplus} and $\partial \bm h(\bm z)/\partial \bm z|_{\bm z = \bm x}$ is calculated as
\begin{equation*}
\frac{\partial \bm{h}\left( \bm{z},\bm{0} \right)}{\partial \bm{z}}= \begin{bmatrix}
	\bm{I}_3&		\bm{0}_{3\times 3}&		\rho _1\bm{I}_3&		\bm{0}_{3\times 3}&		-l_1\bm{I}_3&		\bm{0}_{3\times 3}&		\bm{0}_{3\times 3}&		\bm{0}_{3\times 3}&		\bm{0}_{3\times 6}\\
	\bm{0}_{3\times 3}&		\bm{I}_3&		\rho _1\bm{\omega }_{0}^{\times}&		-\rho _1\bm{q}_{0}^{\times}&		-l_1\bm{\omega }_{1}^{\times}&		l_1\bm{q}_{1}^{\times}&		\bm{0}_{3\times 3}&		\bm{0}_{3\times 3}&		\bm{0}_{3\times 6}\\
	\bm{I}_3&		\bm{0}_{3\times 3}&		-\rho _2\bm{I}_3&		\bm{0}_{3\times 3}&		\bm{0}_{3\times 3}&		\bm{0}_{3\times 3}&		-l_2\bm{I}_3&		\bm{0}_{3\times 3}&		\bm{0}_{3\times 6}\\
	\bm{0}_{3\times 3}&		\bm{I}_3&		-\rho _2\bm{\omega }_{0}^{\times}&		\rho _2\bm{q}_{0}^{\times}&		\bm{0}_{3\times 3}&		\bm{0}_{3\times 3}&		-l_2\bm{\omega }_{2}^{\times}&		l_2\bm{q}_{2}^{\times}&		\bm{0}_{3\times 6}
\end{bmatrix}.
\end{equation*}

\bibliographystyle{IEEEtran}

\bibliography{reference}

@article{klausen2018cooperative,
  title={Cooperative control for multirotors transporting an unknown suspended load under environmental disturbances},
  author={Klausen, Kristian and Meissen, Chris and Fossen, Thor I and Arcak, Murat and Johansen, Tor Arne},
  journal={IEEE Transactions on Control Systems Technology},
  volume={28},
  number={2},
  pages={653--660},
  year={2018},
  publisher={IEEE}
}

@article{lee2017geometric,
  title={Geometric control of quadrotor UAVs transporting a cable-suspended rigid body},
  author={Lee, Taeyoung},
  journal={IEEE Transactions on Control Systems Technology},
  volume={26},
  number={1},
  pages={255--264},
  year={2017},
  publisher={IEEE}
  }

@inproceedings{sreenath2013dynamics,
  title={Dynamics, Control and Planning for Cooperative Manipulation of Payloads Suspended by Cables from Multiple Quadrotor Robots},
  author={Koushil Sreenath and Vijay R. Kumar},
  booktitle={Robotics: Science and Systems},
  year={2013},
  address={Berlin, Germany},
  month={Jun.}
}

@article{zeng2020differential,
  title={Differential flatness based path planning with direct collocation on hybrid modes for a quadrotor with a cable-suspended payload},
  author={Zeng, Jun and Kotaru, Prasanth and Mueller, Mark W and Sreenath, Koushil},
  journal={IEEE Robotics and Automation Letters},
  volume={5},
  number={2},
  pages={3074--3081},
  year={2020},
  publisher={IEEE}
}

@ARTICLE{xu2024force,
  author={Xu, Lidan and Lu, Hao and Wang, Jianliang and Oh, Hyondong and Guo, Xiang-Gui and Guo, Lei},
  journal={IEEE Transactions on Aerospace and Electronic Systems}, 
  title={Force-Coordination Control for Aerial Collaborative Transportation Based on Lumped Disturbance Separation and Estimation}, 
  year={2024},
  volume={60},
  number={5},
  pages={6037-6049},
  doi={10.1109/TAES.2024.3401026}}

@article{klausen2017nonlinear,
  title={Nonlinear Control With Swing Damping of a Multirotor UAV with Suspended Load},
  author={Klausen, Kristian and Fossen, Thor I and Johansen, Tor Arne},
  journal={Journal of Intelligent \& Robotic Systems},
  volume={88},
  pages={379--394},
  year={2017},
  publisher={Springer}
}

@article{geng2020cooperative,
  title={Cooperative Transport of a Slung Load Using Load-Leading Control},
  author={Geng, Junyi and Langelaan, Jack W},
  journal={Journal of Guidance, Control, and Dynamics},
  volume={43},
  number={7},
  pages={1313--1331},
  year={2020},
  publisher={American Institute of Aeronautics and Astronautics}
}

@article{li2021cooperative,
  title={Cooperative Transportation of Cable Suspended Payloads With {MAVs} Using Monocular Vision and Inertial Sensing},
  author={Li, Guanrui and Ge, Rundong and Loianno, Giuseppe},
  journal={IEEE Robotics and Automation Letters},
  volume={6},
  number={3},
  pages={5316--5323},
  year={2021},
  publisher={IEEE}
}

@inproceedings{lee2017autonomous,
  title={Autonomous Swing-Angle Estimation for Stable Slung-Load Flight of Multi-Rotor {UAVs}},
  author={Lee, Seung Jae and Kim, H Jin},
  booktitle={2017 IEEE International Conference on Robotics and Automation (ICRA)},
  pages={4576--4581},
  year={2017},
  organization={IEEE}
}

@article{de2019swing,
  title={Swing angle estimation for multicopter slung load applications},
  author={de Angelis, Emanuele L},
  journal={Aerospace Science and Technology},
  volume={89},
  pages={264--274},
  year={2019},
  publisher={Elsevier}
}

@article{liang2021fault,
  title={Fault-tolerant control for the multi-quadrotors cooperative transportation under suspension failures},
  author={Liang, Xiao and Su, Zikang and Zhou, Wenzhe and Meng, Guanglei and Zhu, Linlin},
  journal={Aerospace Science and Technology},
  volume={119},
  pages={107139},
  year={2021},
  publisher={Elsevier}
}

@article{wang2022geometric,
  title={Geometric control for trajectory-tracking of a quadrotor {UAV} with suspended load},
  author={Wang, Junan and Yuan, Xiaozhuoer and Zhu, Bing},
  journal={IET Control Theory \& Applications},
  volume={16},
  number={12},
  pages={1271--1281},
  year={2022},
  publisher={Wiley Online Library}
}

@article{pereira2020pose,
  title={Pose stabilization of a bar tethered to two aerial vehicles},
  author={Pereira, Pedro O and Dimarogonas, Dimos V},
  journal={Automatica},
  volume={112},
  pages={108695},
  year={2020},
  publisher={Elsevier}
}

@article{erskine2019wrench,
  title={Wrench analysis of cable-suspended parallel robots actuated by quadrotor unmanned aerial vehicles},
  author={Erskine, Julian and Chriette, Abdelhamid and Caro, St{\'e}phane},
  journal={Journal of Mechanisms and Robotics},
  volume={11},
  number={2},
  pages={020909},
  year={2019},
  publisher={American Society of Mechanical Engineers}
}

@article{xu2024oscillation,
  title={Oscillation Suppression-Enhanced Cooperative Control via Refined Cooperative Disturbance Estimation for Aerial Co-Transportation System},
  author={Xu, Lidan and Lu, Hao and Wang, Jianliang and Oh, Hyondong and Guo, Xiang-Gui and Guo, Lei},
  journal={IEEE Transactions on Automation Science and Engineering},
  year={2024},
  publisher={IEEE}
}

@ARTICLE{payloadIMU,
  author={Zhang, Zhuang and Yu, Hai and Ye, Huiying and Han, Jianda and Fang, Yongchun and Liang, Xiao},
  journal={IEEE Transactions on Industrial Informatics}, 
  title={Collaborative Control for Aerial Transportation of Cargo With Dual Quadrotors}, 
  year={2025},
  volume={21},
  number={1},
  pages={752-761},
  doi={10.1109/TII.2024.3459077}}

@article{GPSRTK,
  title={Outdoor swarm flight system based on rtk-gps},
  author={Moon, SungTae and Choi, YeonJu and Kim, DoYoon and Seung, Myeonghun and Gong, HyeonCheol},
  journal={Journal of KIISE},
  volume={43},
  number={12},
  pages={1315--1324},
  year={2016},
  publisher={Korean Institute of Information Scientists and Engineers}
}

@article{d2slam,
  title={${D}^{2}${SLAM}: Decentralized and Distributed Collaborative Visual-Inertial SLAM System for Aerial Swarm},
  author={Xu, Hao and Liu, Peize and Chen, Xinyi and Shen, Shaojie},
  journal={IEEE Transactions on Robotics},
  year={2024},
  publisher={IEEE}
}

@article{swarmlio2,
  title={Swarm-{LIO}2: Decentralized, Efficient LiDAR-inertial Odometry for UAV Swarms},
  author={Zhu, Fangcheng and Ren, Yunfan and Yin, Longji and Kong, Fanze and Liu, Qingbo and Xue, Ruize and Liu, Wenyi and Cai, Yixi and Lu, Guozheng and Li, Haotian and others},
  journal={IEEE Transactions on Robotics},
  year={2024},
  publisher={IEEE}
}

@article{de2023improved,
  title={An improved method for swing state estimation in multirotor slung load applications},
  author={de Angelis, Emanuele Luigi and Giulietti, Fabrizio},
  journal={Drones},
  volume={7},
  number={11},
  pages={654},
  year={2023},
  publisher={MDPI}
}

@inproceedings{mellet2023neural,
  title={Neural-Network for Position Estimation of a Cable-Suspended Payload Using Inertial Quadrotor Sensing},
  author={Mellet, Julien and Cacace, Jonathan and Ruggiero, Fabio and Lippiello, Vincenzo and others},
  booktitle={ICINCO (1)},
  pages={80--87},
  year={2023}
}

@ARTICLE{xie2020,
  author={Xie, Heng and Dong, Kaixu and Chirarattananon, Pakpong},
  journal={IEEE Access}, 
  title={Cooperative Transport of a Suspended Payload via Two Aerial Robots With Inertial Sensing}, 
  year={2022},
  volume={10},
  number={},
  pages={81764-81776},
  doi={10.1109/ACCESS.2022.3194932}}

@article{goodman2022geometric,
  title={Geometric control of two quadrotors carrying a rigid rod with elastic cables},
  author={Goodman, Jacob and Colombo, Leonardo},
  journal={Journal of Nonlinear Science},
  volume={32},
  number={5},
  pages={65},
  year={2022},
  publisher={Springer}
}

@ARTICLE{arthur,
  author={Hermann, R. and Krener, A.},
  journal={IEEE Transactions on Automatic Control}, 
  title={Nonlinear controllability and observability}, 
  year={1977},
  volume={22},
  number={5},
  pages={728-740},
  doi={10.1109/TAC.1977.1101601}}

@book{martinelli2020observability,
  title={Observability: A New Theory Based on the Group of Invariance},
  author={Martinelli, Agostino},
  year={2020},
  publisher={SIAM}
}

@article{he2023symbolic,
  title={Symbolic representation and toolkit development of iterated error-state extended kalman filters on manifolds},
  author={He, Dongjiao and Xu, Wei and Zhang, Fu},
  journal={IEEE Transactions on Industrial Electronics},
  volume={70},
  number={12},
  pages={12533--12544},
  year={2023},
  publisher={IEEE}
}

@article{hertzberg2013integrating,
  title={Integrating generic sensor fusion algorithms with sound state representations through encapsulation of manifolds},
  author={Hertzberg, Christoph and Wagner, Ren{\'e} and Frese, Udo and Schr{\"o}der, Lutz},
  journal={Information Fusion},
  volume={14},
  number={1},
  pages={57--77},
  year={2013},
  publisher={Elsevier}
}

@article{markley2003attitude,
  title={Attitude error representations for Kalman filtering},
  author={Markley, F Landis},
  journal={Journal of Guidance, Control, and Dynamics},
  volume={26},
  number={2},
  pages={311--317},
  year={2003}
}

@article{mirzaei2008kalman,
  title={A Kalman filter-based algorithm for IMU-camera calibration: Observability analysis and performance evaluation},
  author={Mirzaei, Faraz M and Roumeliotis, Stergios I},
  journal={IEEE Transactions on Robotics},
  volume={24},
  number={5},
  pages={1143--1156},
  year={2008},
  publisher={IEEE}
}

@article{sola2017quaternion,
  title={Quaternion kinematics for the error-state Kalman filter},
  author={Sola, Joan},
  journal={arXiv preprint arXiv:1711.02508},
  year={2017}
}

@article{li2013high,
  title={High-precision, consistent EKF-based visual-inertial odometry},
  author={Li, Mingyang and Mourikis, Anastasios I},
  journal={The International Journal of Robotics Research},
  volume={32},
  number={6},
  pages={690--711},
  year={2013},
  publisher={Sage Publications Sage UK: London, England}
}

@article{xu2021fast,
  title={Fast-lio: A fast, robust lidar-inertial odometry package by tightly-coupled iterated kalman filter},
  author={Xu, Wei and Zhang, Fu},
  journal={IEEE Robotics and Automation Letters},
  volume={6},
  number={2},
  pages={3317--3324},
  year={2021},
  publisher={IEEE}
}

@inproceedings{px4ekf,
  title={Research on the hardware structure characteristics and EKF filtering algorithm of the autopilot PIXHAWK},
  author={Feng, Lin and Fangchao, Qi},
  booktitle={2016 Sixth International Conference on Instrumentation \& Measurement, Computer, Communication and Control (IMCCC)},
  pages={228--231},
  year={2016},
  organization={IEEE}
}

@article{guo2023composite,
  title={Composite disturbance filtering: A novel state estimation scheme for systems with multisource, heterogeneous, and isomeric disturbances},
  author={Guo, Lei and Li, Wenshuo and Zhu, Yukai and Yu, Xiang and Wang, Zidong},
  journal={IEEE Open Journal of the Industrial Electronics Society},
  volume={4},
  pages={387--400},
  year={2023},
  publisher={IEEE}
}

@book{guo2013anti,
  title={Anti-disturbance control for systems with multiple disturbances},
  author={Guo, Lei and Cao, Songyin},
  year={2013},
  publisher={CRC Press}
}

@article{guo2014anti,
  title={Anti-disturbance control theory for systems with multiple disturbances: A survey},
  author={Guo, Lei and Cao, Songyin},
  journal={ISA Transactions},
  volume={53},
  number={4},
  pages={846--849},
  year={2014},
  publisher={Elsevier}
}

@article{guo2012initial,
  title={Initial alignment for nonlinear inertial navigation systems with multiple disturbances based on enhanced anti-disturbance filtering},
  author={Guo, Lei and Cao, Songyin and Qi, Chuntang and Gao, Xiaoying},
  journal={International Journal of Control},
  volume={85},
  number={5},
  pages={491--501},
  year={2012},
  publisher={Taylor \& Francis}
}

\end{document}